\documentclass[lettersize,journal]{IEEEtran}
\usepackage{amsmath,amsfonts,amssymb}
\usepackage{array}
\usepackage{textcomp}
\usepackage{stfloats}
\usepackage{url}
\usepackage{verbatim}
\usepackage{graphicx}
\usepackage{multirow}
\usepackage{tensor}
\usepackage{siunitx}
\usepackage{mathrsfs}
\usepackage{booktabs}
\usepackage{color}
\usepackage{pifont}
\usepackage{subcaption} 
\usepackage{algorithm}  
\usepackage{algpseudocode} 
\usepackage{tabularx}
\usepackage{gensymb}
\usepackage{lipsum}
\usepackage{arydshln} 

\usepackage{hyperref}
\hypersetup{colorlinks,breaklinks,
	linkcolor=[rgb]{0.5,0.,0.},
	citecolor=[rgb]{0.000,0.25,0.05},
	urlcolor=[rgb]{0.031,0.318,0.612}}

\usepackage[table]{xcolor}
\definecolor{myOrange}{rgb}{1, 0.5, 0.055} 
\definecolor{myRed}{rgb}{0.84, 0.15, 0.157} 
\definecolor{myGreen}{rgb}{0.173, 0.627, 0.173} %
\definecolor{top1}{RGB}{111, 170, 247}
\definecolor{top2}{RGB}{169, 204, 250}
\definecolor{top3}{RGB}{226, 238, 253}

\newcommand{\halfcmark}{\checkmark\kern-1.1ex\raisebox{.7ex}{\rotatebox[origin=c]{125}{--}}}

\def\secref#1{Sec.~\ref{#1}}
\def\figref#1{Fig.~\ref{#1}}
\def\tabref#1{Tab.~\ref{#1}}
\def\eqref#1{Eq.~(\ref{#1})}

\def\BibTeX{{\rm B\kern-.05em{\sc i\kern-.025em b}\kern-.08em
    T\kern-.1667em\lower.7ex\hbox{E}\kern-.125emX}}
\usepackage{balance}

\begin{document}
\title{SiLVR: Scalable Lidar-Visual Radiance Field Reconstruction with Uncertainty Quantification}

\author{Yifu Tao$^1$, Maurice Fallon$^1$}

\markboth{IEEE TRANSACTIONS ON ROBOTICS,~Vol.xx, No.xx, Month ~Year}
{Tao \MakeLowercase{\textit{et al.}}: SiLVR: Uncertainty-Aware Lidar-Visual NeRF} 

\twocolumn[{%
\renewcommand\twocolumn[1][]{#1}%
\maketitle
\begin{center}
    \vspace{-25pt}
    \centering
	\includegraphics[width=1\textwidth]{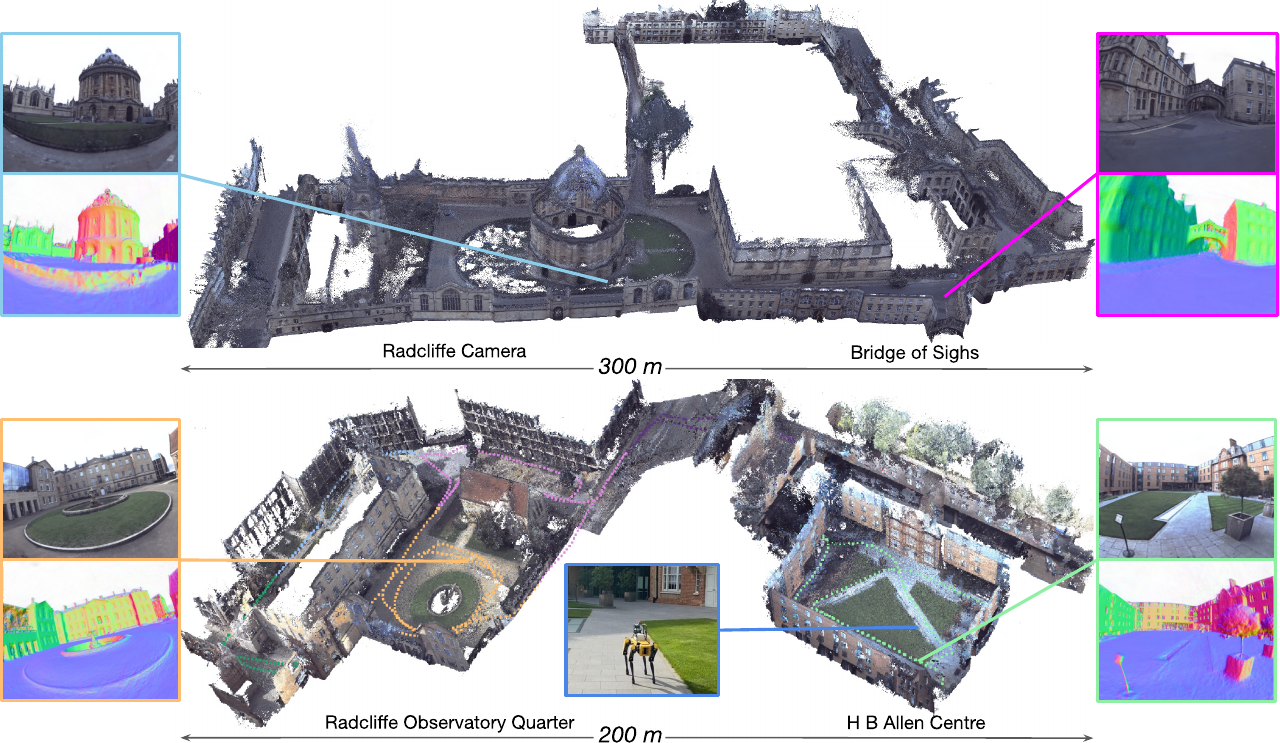}
	\captionof{figure}{Two large-scale reconstructions generated by SiLVR. Rendered RGB and surface normal images from the reconstructions are shown on each side. SiLVR combines visual and lidar information to create geometrically accurate maps with photorealistic textures, while considering sensor uncertainty. SiLVR uses submaps to scale to large-scale building complexes.} 
    \label{fig:hero}
\end{center}%
}]

\begingroup
  \renewcommand\thefootnote{}\footnote{\hspace{-1em}
  $^1$Oxford Robotics Inst., Dept. of Engineering Sci., Univ. of Oxford, UK. \\
 This project has been partly funded by the Horizon Europe project Digiforest (101070405) and the National Research Foundation of Korea (NRF) grant funded by the Korea government (MSIT)(No. RS-2024-00461409). Maurice Fallon is supported by a Royal Society University Research Fellowship. For the purpose of open access, the authors have applied a Creative Commons Attribution (CC BY) license to any Accepted Manuscript version arising.
  }
  \addtocounter{footnote}{-1}%
\endgroup

\begin{abstract}
We present a neural radiance field (NeRF) based large-scale reconstruction system that fuses lidar and vision data to generate high-quality reconstructions that are geometrically accurate and capture photorealistic texture. Our system adopts the state-of-the-art NeRF representation to additionally incorporate lidar. Adding lidar data adds strong geometric constraints on the depth and surface normals, which is particularly useful when modelling uniform texture surfaces which contain ambiguous visual reconstruction cues. 
A key contribution of this work is a novel method to quantify the epistemic uncertainty of the lidar-visual NeRF reconstruction by estimating the spatial variance of each point location in the radiance field given the sensor observations from the cameras and lidar. This provides a principled approach to evaluate the contribution of each sensor modality to the final reconstruction. In this way, reconstructions that are uncertain (due to, e.g., uniform visual texture, limited observation viewpoints, or little lidar coverage) can be identified and removed. Our system is integrated with a real-time pose-graph lidar SLAM system which is used to bootstrap a Structure-from-Motion (SfM)
reconstruction procedure. It also helps to properly constrain the overall metric scale which is essential for the lidar depth loss. The refined SLAM trajectory can then be divided into submaps using Spectral Clustering to group sets of co-visible images together. This submapping approach is more suitable for visual reconstruction than distance-based partitioning. Our uncertainty estimation is particularly effective when merging submaps, as their boundaries often contain artefacts due to limited observations. We demonstrate the reconstruction system using a multi-camera, lidar sensor suite in experiments involving both robot-mounted and handheld scanning. Our test datasets cover a total area of more than \( \text{20,000}~\text{m}^2 \), including multiple university buildings and an aerial survey of a multi-storey building. Quantitative evaluation is provided by comparing with maps produced by a commercial tripod scanner. The code and dataset are available at \url{https://dynamic.robots.ox.ac.uk/projects/silvr/}.

\end{abstract}

\section{Introduction}

Dense 3D reconstruction is a core component of a range of robotics applications such as industrial inspection and autonomous navigation. Common sensors used for reconstruction include cameras and lidars. Camera-based reconstruction systems use techniques including Structure-from-Motion (SfM)~\cite{schoenberger2016colmap} and Multi-View Stereo (MVS)~\cite{schoenberger2016mvs} to produce dense textured reconstructions. However, these approaches rely on good lighting conditions and can require exhaustive data collection to capture data from diverse viewpoints. They also struggle with textureless areas such as bare walls, ceilings and floors. A lidar sensor provides accurate geometric information at long range---as it actively measures distances to surfaces. This makes lidar desirable for large-scale outdoor environments. However, lidar scans are much sparser than camera images. They also do not capture colour which is important for applications such as virtual reality and 3D asset generation.

Classical reconstruction systems have used point clouds, occupancy maps, and sign-distance fields (SDF) as their internal 3D representation. Recently, radiance field representations, namely neural radiance fields (NeRF)~\cite{mildenhall2021nerf} and 3D Gaussian Splatting (3DGS)~\cite{kerbl3Dgaussians} have gained popularity for visual reconstruction. Taking advantage of differentiable rendering, these techniques optimise a 3D representation by minimising the difference between a rendered image and a reference camera image. These methods can synthesise novel views with near photorealistic quality, which can be useful for robotic inspection and visual localisation.

As with traditional vision-based reconstruction methods, NeRF struggles to estimate accurate geometry in locations where there is limited multi-view input (i.e., when images are only taken from a single direction) or sparse texture. Autonomous systems commonly encounter these situations---for example, an inspection robot moving only in a forward direction might only obtain visual observations from parallel viewpoints, which makes triangulation of visual features more difficult. Additionally, a robot operating in man-made environments often encounters objects such as textureless walls which are difficult to reconstruct using only vision. As a result, radiance field reconstruction of feature-less regions (e.g., the ground reconstruction using Nerfacto in \figref{fig:reconstruction_eval_large_scale}; the sky and clouds reconstructed as floating points on the top of the scene~\cite{rematas2022urban}) and objects with limited viewpoints (e.g., reconstruction of a wall viewed from a single direction shown in \figref{fig:mono_3cam}) are often inaccurate. These challenges affect not only volume-density-based fields such as NeRFs, but also SDF-based radiance fields such as NeuS~\cite{wang2021neus} and 3D Gaussian representations~\cite{kerbl3Dgaussians} (e.g., NeuSfacto reconstruction in \figref{fig:reconstruction_eval_large_scale} and elongated Gaussians described in MonoGS~\cite{matsuki2024monogs}). In addition, the implicit representation used in NeRF, while providing tremendous model size compression compared to explicit representations such as 3DGS~\cite{kerbl3Dgaussians}, can generate reconstruction artefacts in unobserved regions of space. This is due to NeRF formulation as a continuous volumetric field over $\mathbb{R}^3$, which enables it to produce outputs even in areas lacking observation. Unlike classical visual SLAM/MVS methods (such as MonoSLAM~\cite{davison2007monoslam} and OpenMVS) which estimate the uncertainties for the visual reconstructions to tackle reconstruction artefacts, the NeRF representation itself has no notion of uncertainties. These factors limit the use of radiance fields in real-world robotic applications where 3D reconstructions have to be reliable.

The challenges of 3D reconstruction using only visual sensing can be mitigated by fusing range sensing devices such as lidar which provides accurate geometric measurements regardless of object texture. Several recent reconstruction systems~\cite{tao2024silvr,zhao2025lvigs,hong2025gslivo} have adopted a lidar-visual radiance field approach, since both RGB colour and depth measurements can be obtained (in a differentiable manner) from the radiance field and then supervised by lidar and camera data. However, lidar measurements are sparse, and have limited range and field-of-view (FOV). As a result, not all parts of the scene can be scanned by a lidar sensor. For example, the lidar in \figref{fig:lidar_visual_unc} has a limited FOV, and the top of the buildings are not scanned by it. As another example, the ground shown in the bottom left of \figref{fig:sample_overlay} is beyond the lidar's sensing range. For these regions, a lidar-visual radiance field reconstruction relies mainly on visual data, and the accuracy then depends on the conditioning of visual reconstruction as discussed in the previous paragraph. Because of this, it is crucial to properly quantify the individual contributions of visual and lidar information in the resultant lidar-visual radiance field reconstruction. However, this issue is relatively under-explored in the literature.

The reconstruction task becomes even more challenging when the scene is large-scale (e.g., urban districts). It is time-consuming to estimate sensor trajectories by running an incremental SfM algorithm such as COLMAP~\cite{schoenberger2016colmap} for a large-scale scene. In addition, SfM is not guaranteed to succeed in registering all input images (especially if parts of the scene have poor lighting), which could then lead to an unreconstructed region in the map. As the scene and the dataset size grow, the model capacity of a NeRF and memory constraints become a bottleneck. Simply increasing the size of the model parameters (e.g., hash table size in Instant-NGP~\cite{mueller2022instant} or the number of 3D Gaussians~\cite{kerbl3Dgaussians}) can exceed available computer memory when working on larger scenes. This motivates the development of a scene partitioning strategy. Some existing methods require manually partitioning using heuristics~\cite{tancik2022block} or evenly partitioning the scene using a grid~\cite{meganerf}. The limitation of a simple grid-based partition is that the view orientation and visibility are not considered. These factors are, however, important to consider when carrying out clustering for visual reconstruction. For example, an image taken from a corridor outside of a room but looking into it ought to be considered part of a submap of that room.

In this work, we present SiLVR, a submap-based NeRF reconstruction system that integrates both lidar and visual information to produce accurate, textured, and uncertainty-aware 3D reconstructions which can also synthesise photorealistic novel views. SiLVR builds upon existing NeRF research and the Nerfacto implementation~\cite{nerfstudio} 
which utilises hash encoding~\cite{mueller2022instant} that is significantly faster than MLP-based NeRF~\cite{mildenhall2021nerf} (it takes less than 5 minutes to train a NeRF for one submap in our experiments). We extend this work by adding geometric constraints from lidar to improve reconstruction quality. Our use of lidar data is particularly important for modelling featureless areas where geometry cannot be accurately reconstructed using 3D SfM features~\cite{deng2022depth}. In addition, we also estimate surface normals from the lidar scans to encourage smooth surface reconstruction. This approach does not suffer from input data distribution shift, a characteristic of learning-based normal estimation approaches~\cite{Yu2022MonoSDF}.

Compared to prior NeRF-based reconstruction systems that incorporate lidar~\cite{tao2024silvr} or depth cameras~\cite{azinovic2022neural,Yu2022MonoSDF}, our key innovation is a rigorous study of how to quantify uncertainty in the resultant reconstruction, which enables improved reconstruction accuracy as well as facilitating downstream tasks such as view selection~\cite{strong2024next} and navigation within a NeRF map~\cite{adamkiewicz2022vision}. After training, SiLVR computes the epistemic uncertainty of the NeRF with the Laplace approximation (LA)~\cite{daxberger2021laplace} and Fisher-Information-approximated Hessian matrix. As an efficient alternative to ensemble learning (which requires multiple iterations of training of the NeRF model and is time-consuming), we build upon the formulation of the perturbation field proposed in BayesRays~\cite{goli2023bayesrays} and use the spatial variance of the field as the measure of epistemic uncertainty. The estimated uncertainty is used to filter out reconstruction artefacts which improves the final reconstruction accuracy. This is particularly important when merging NeRF submaps since these submaps often contain artefacts caused by limited observation at the submap boundaries. In addition, we adapt a previously developed lidar SLAM algorithm~\cite{wisth2023vilens,ramezani2020slam} to bootstrap the SfM component and to enforce an accurate metric scale. For mapping large-scale building complexes or a city block, we adopt a submapping approach and apply a graph partitioning algorithm~\cite{shi2000normalized} with visibility information to divide the complete trajectory into submaps. We study how the use of visibility information allows the submaps to be more self-contained and to have fewer artefacts at their boundaries compared to methods that only consider spatial information~\cite{tao2024silvr}. 

In summary, our contributions are as follows:
\begin{itemize}
	\item An uncertainty-aware NeRF reconstruction system fusing lidar and vision that can reconstruct large-scale outdoor environments.
    \item Epistemic uncertainty quantification of the multi-modal NeRF pipeline as a principled approach to quantify the contribution of lidar and visual information to a 3D reconstruction, which can be used to improve reconstruction accuracy, especially at submap boundaries. Our method extends the vision-only uncertainty estimation framework proposed in BayesRays~\cite{goli2023bayesrays} to also support lidar depth. This allows us to identify areas with reliable reconstruction (e.g., where there are visual features or abundant lidar measurements) and unreliable areas (e.g., uniformly-textured surfaces which have also not been scanned by lidar). 
    \item A submapping strategy that leverages per-image visibility information. Compared to the distance-based clustering method~\cite{tao2024silvr}, we develop a visibility-based clustering method which reduces visual overlap between submaps and in turn creates fewer artefacts at submap boundaries.
    \item Large-scale evaluation using two large-scale datasets from the Oxford Spires dataset~\cite{tao2025spires} with quantitative results from millimetre-accurate 3D ground truth. Further comparison is made to baseline radiance field methods that use SDF~\cite{wang2021neus} and 3D Gaussians~\cite{kerbl3Dgaussians} representations.
\end{itemize}

\section{Related Work}
\label{sec:related_works}

In this section, we first review classical 3D reconstruction methods based on lidar sensors or cameras, followed by more recent approaches using radiance field representations. We then discuss methods to quantify the uncertainty of a NeRF reconstruction, and techniques that can extend the NeRF methods to large-scale environments.

\subsection{Classical 3D Reconstruction}
Lidars and cameras are the two main modalities used in robotic perception and specifically for 3D reconstruction. For each sensor modality, trajectory estimation is a typical first step in a reconstruction pipeline. In this section, we review classical lidar-based and vision-based pose estimation and 3D reconstruction methods. Then, we review strategies for extending these methods to large-scale scenes.

Lidar is the dominant sensor used for accurate 3D reconstruction of large-scale outdoor environments~\cite{behley2018rss, lin2022r3live}. It actively transmits laser pulses to measure ranges and, as a result, is accurate even at ranges beyond 100m. With these distance measurements, Lidar odometry typically uses a variant of Iterative Closest Point (ICP), and often integrates high-frequency IMU measurements~\cite{zhang2014loam, zhao2021super, xu2022fast, wisth2023vilens}. Small errors in odometry can accumulate over time, resulting in trajectory drift. This drift can be mitigated when a sensor revisits a previous place and detects loop closure with pose graph optimisation~\cite{thrun2006graph} which allows a consistent map to be maintained.
After registering all the lidar scans, the (surface) reconstruction problem is then to fuse discrete observations into a map. 
Example map representations include surfels~\cite{whelan2016elasticfusion,park2018elastic}, voxels~\cite{hornung13octomap,newcombe2011kinectfusion,oleynikova2017voxblox,Vespa2018supereight} and wavelets~\cite{reijgwart2023wavemap}. 
Despite its advantages, lidar has its own limitations. Lidar sensors are much more expensive than cameras, and their measurements are typically much sparser than camera images. The measurements have inherent noise with ranging errors in the order of centimetres, which makes it difficult to reconstruct small objects accurately in indoor scenes. Finally, lidar data contains no texture or colour, so the final reconstruction is only geometric and cannot be used for applications such as view synthesis, which requires texture.

Alternatively, textured reconstructions can be recovered from camera images alone via SfM. Given the correspondences between images, a SfM system~\cite{schoenberger2016colmap} can optimise a set of camera poses, camera intrinsics, and 3D sparse feature points. This can then be used by a MVS system~\cite{furukawa2010dense} to compute dense depth for each frame and in turn to create a dense 3D point cloud. Compared to lidar, cameras are much more affordable, and also provide texture and colour. However, the performance of visual reconstruction method depends on environmental conditions, and the quality of feature matching. This makes the resultant reconstruction less reliable in scenes that contain repetitive patterns, low-texture surfaces, poor lighting conditions and non-Lambertian materials.



When the scene to be reconstructed is large-scale (e.g., urban districts or multi-room indoor environments),  computer memory becomes a limiting factor. Attempting to map a large scene while constraining the output map size might result in a lower-resolution reconstruction lacking detail. A common strategy to extend dense reconstructions to large-scale areas is to divide the scene into submaps~\cite{bosse2003atlas}. For visual reconstruction with many thousands of images~\cite{agarwal2011building}, a submapping approach can significantly reduce computation time and memory requirements. One approach used in large-scale MVS is the submap partitioning~\cite{furukawa2010towards} which groups images into clusters while not degrading the final resultant reconstruction. After partitioning, each submap should be filtered and merged into one unified model. For online lidar mapping systems, the motivation for using submapping techniques is to accommodate loop closure corrections into the already-built map (Occupancy grid or TSDF). These systems construct submaps that are attached to a pose graph~\cite{ho2018virtual,reijgwart2020voxgraph,wang2022strategies}, and can deform each submap by reoptimising the pose graph upon loop closure.

\begin{figure*}[t]
	\centering
	\includegraphics[width=2\columnwidth]{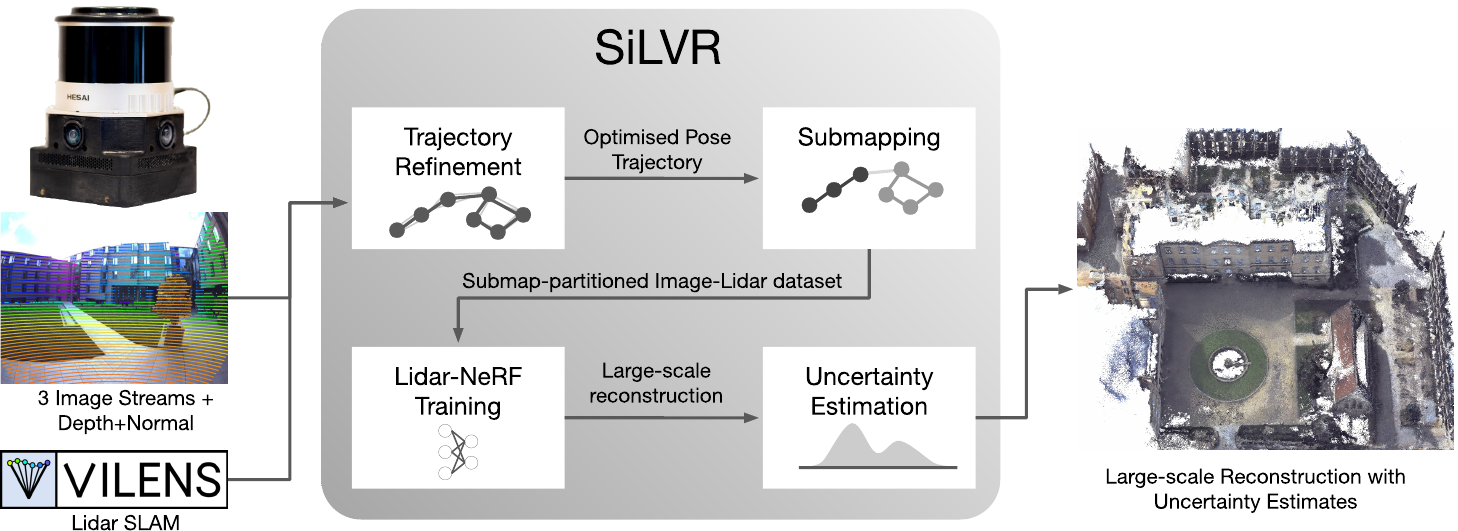}
	\caption{System overview: SiLVR builds large-scale reconstructions using images and lidar data, and a pose trajectory estimated by a separate odometry system. The sensor streams are provided by the \emph{Frontier}, our custom perception payload carrying three fisheye colour cameras, IMU measurements, and a 3D lidar. When collecting the data, we used VILENS~\cite{wisth2023vilens} to estimate the trajectory of the sensor, which is refined in post-processing using COLMAP~\cite{schoenberger2016colmap} and partitioned into submaps. The camera image, lidar depth, and a derivative normal image are used to train a NeRF to achieve a final 3D reconstruction. After training the NeRF, SiLVR estimates the epistemic uncertainty of the radiance field. Finally, the point cloud reconstruction is extracted from the NeRF by rendering a depth for each of the training rays. The point cloud is then filtered using per-point uncertainty estimates to remove unreliable reconstructions.}
	\label{fig:sys_overview}
\end{figure*}

\subsection{Radiance Field Representation}
Neural Radiance Fields (NeRF) were first proposed in the seminal paper from Mildenhall et al.~\cite{mildenhall2021nerf}. The technique uses a multilayer perceptron (MLP) to represent a continuous radiance field, and uses differentiable volume rendering to reconstruct novel views. It minimises the photometric error between the rendered image and the input image, which implicitly achieves multi-view consistency.
NeRF and its many variants use frequency encoding~\cite{vaswani2017attention} to encode spatial coordinates, but these suffer from long training times, typically a few hours per scene. Alternative explicit representations of radiance fields, including dense voxel-grids with trainable per-vertex features~\cite{fridovich2022plenoxels,mueller2022instant} and more recently 3D Gaussians~\cite{kerbl3Dgaussians}, are shown to accelerate the training, at the cost of being more memory intensive.
Octree or sparse-grid structures~\cite{yu2021plenoctrees,mueller2022instant} can reduce memory usage by pruning grid-features in empty space. 
Our work is built upon Nerfacto from the open-sourced Nerfstudio project~\cite{nerfstudio}. It incorporates the main features from other NeRF works~\cite{mueller2022instant,barron2022mipnerf360,martinbrualla2020nerfw} that have been found to work well with real data.


While NeRFs excel at high-quality view synthesis, obtaining a 3D surface reconstruction of similar quality remains challenging, mainly due to the flexible volumetric representation being under-constrained by the limited multi-view inputs. One approach to improve the reconstruction is to impose depth regularisation~\cite{deng2022depth, rematas2022urban} or surface normal regularisation~\cite{Yu2022MonoSDF} which can be obtained from depth sensors or be estimated by a neural network.
Another approach is to impose surface priors on the volumetric field and use 
representations such as Signed Distance Field (SDF)~\cite{neuralangelo,yariv2021volume} and 2D Gaussians~\cite{huang20242dgs} to enforce a surface reconstruction output, although the novel view synthesis quality might be compromised~\cite{wang2021neus} with this approach. 
Our method uses a volume density representation which is extended with depth~\cite{deng2022depth} and surface normal~\cite{Yu2022MonoSDF} regularisations from lidar measurements instead of using SfM~\cite{deng2022depth} or learnt priors~\cite{Yu2022MonoSDF}. This can significantly improve the reconstruction quality in texture-less areas with smooth surfaces.

Neural field representations have been used for lidar-based mapping~\cite{zhong2023icra,deng2023nerfloam,pan2024pinslam}, showing promise in generating more complete and compact reconstructions than traditional methods. While these works also build upon implicit map representations, they do not use visual data to build the map. Our system uses visual information and multi-view geometry constraints. Because of this, it can reconstruct regions outside of the lidar's FOV.

\subsection{Uncertainty in Neural Radiance Fields}
The standard formulation of NeRF has no notion of uncertainty. The lack of uncertainty makes it difficult to apply them in robotics applications because a NeRF reconstruction could contain artefacts. From a Bayesian machine learning perspective, one could model the data uncertainty (aleatoric uncertainty) and model uncertainty (epistemic uncertainty)~\cite{kendall2017uncertainties} in the NeRF model. The data uncertainty models how the image observation differs from the trained NeRF, and the source of errors includes dynamic objects, lighting conditions and non-Lambertian surfaces. Dynamic object masking~\cite{Sabour2023robustnerf} and appearance encoding~\cite{martinbrualla2020nerfw} have been used to model or mitigate data uncertainty.

The model uncertainty aims to capture the variance of the radiance field given the training data. For example, for a uniformly-textured area (e.g., sky) with parallel viewing angles, there are infinite possible NeRF solutions that can lead to exactly the same image pixel observation. In comparison, the NeRF of a textured object with a clear boundary and observations from multiple viewpoints would have low model uncertainty, and this is similar to the well-conditioned scenario for photogrammetry. The most straightforward and reliable way to quantify model uncertainty is to train an ensemble of models with different initialisations~\cite{lakshminarayanan2017simple}. BayesRays~\cite{goli2023bayesrays} proposes to model the uncertainty of a perturbation field instead, and estimates the uncertainty with the Laplace approximation. We extend BayesRays's perturbation field formulation to also incorporate lidar data, which allows us to obtain uncertainty estimates for both sensor modalities, and filter the results reconstruction considering each sensor's own characteristics.


\subsection{Large-scale Neural Radiance Fields}

Submapping has been used in NeRF representations for city-scale reconstruction. There are several partitioning strategies that are based on grid partitioning~\cite{meganerf} or using road intersections~\cite{tancik2022block}. Merging NeRF submaps is difficult, since each NeRF submap's boundary can be ambiguous, and the appearance encoding of each submap can be different~\cite{martinbrualla2020nerfw}. 
Block-NeRF~\cite{tancik2022block} merges submaps by first selecting submap candidates based on distance and visibility, and combines submaps in the 2D image space with interpolation and test-time appearance matching. These methods either require manual submap partitioning~\cite{tancik2022block}, or partition the scene into 2D grids~\cite{meganerf}. Our work adopts the submapping approach, and develops partitioning strategies based on visibility, which is advantageous compared to 2D grid partitioning of image data that are close in Euclidean distance but in fact belong to isolated regions (e.g., rooms). We also develop novel strategies for submap merging which uses epistemic uncertainty estimation. 





\section{Preliminaries}

\subsection{Radiance Field Formulation}
We start by adopting the radiance field representation and the differentiable volume rendering framework originally proposed by Mildenhall et. al~\cite{mildenhall2021nerf}. The radiance field models the scene as a function $R:(\mathbf{p},\mathbf{d}) \mapsto (\mathbf{c},\sigma)$ where the input includes a 3D location $\mathbf{p}=(p_x,p_y,p_z)$ and 2D viewing direction $\mathbf{d}=(\phi,\psi)$, the output is an emitted colour $\mathbf{c}=(r,g,b)$ and a volume density $\sigma$.
To render a novel view using a NeRF from a particular viewpoint, we cast rays $\mathbf{r}(t)=\mathbf{o}+t\mathbf{d}$ from the camera origin $\mathbf{o}$ along the viewing direction $\mathbf{d}$ of each pixel $\mathbf{u}$ in the image plane, and render the pixel-colour by integrating over the set of points sampled along the ray. The pixel colour $\hat{\mathbf{C}}(\mathbf{r})$ is computed by the volume rendering integral which is approximated using the quadrature rule~\cite{max1995optical,kajiya1984ray} as 
\begin{equation}
\hat{\mathbf{C}}(\mathbf{r}) = \sum_{i=0}^{N} w_i \mathbf{c}_{i},
\label{eq:pixel_colour}
\end{equation}

where $\mathbf{c}_i$ is the colour of the $i$-th point sample along the ray $\mathbf{p}_i$, and its weighting coefficient $w_i$ is defined by 
\begin{equation} \label{eq:rendering}
w_i =   \exp{\left(-\sum_{j=1}^{i-1} \delta_j \sigma_j\right)} (1-\exp{(-\delta_i\sigma_i)}).
\end{equation}
where $\delta_i$ is the distance between adjacent samples, and $\sigma_i$ is the volume density of $\mathbf{p}_i$.

The radiance field can be trained with a squared photometric loss given the ground truth colour $\mathbf{C}(\mathbf{r})$ from the input image:
\begin{equation}
\mathcal{L}_{\text {Colour}}=\sum_{\mathbf{r} \in \mathcal{D}} ||\hat{\mathbf{C}}(\mathbf{r})-\mathbf{C}(\mathbf{r})||^2  
\label{eq:photometric_loss}
\end{equation}
where $\mathcal{D}$ is the whole training dataset used to generate the rays $\mathbf{r}$ and ground truth colour $\mathbf{C}(\mathbf{r})$.

\subsection{Bayesian Interpretation of the NeRF training}
\label{sec:bayesian_nerf}
The NeRF reconstruction (i.e., the radiance field function $f$) is deterministic and has no explicit notion of uncertainty. In practice, different parts of the NeRF reconstruction are inherently more uncertain or less reliable. For example, ill-conditioned visual constraints from non-textured areas or insufficient visual parallax can cause the visual reconstruction accuracy to deteriorate. Another example specifically relevant to our work is that a surface is more uncertain if it has only been sparsely scanned by the lidar with limited range and FOV compared to a surface that is densely scanned. Quantifying these uncertainties associated with the NeRF reconstruction allows one to identify the unreliable reconstructions and filter them out accordingly, hence improving reconstruction accuracy. It can also enable downstream tasks, such as view selection for more accurate and complete mapping.

The Bayesian probability theory provides useful tools for quantifying the uncertainties in neural models~\cite{kendall2017uncertainties}, which can benefit the NeRF reconstruction. For a regression task with the dataset $\mathcal{D}=\{(x_n, y_n)\}_{n=1}$ (where $x_n,y_n$ are the $n$-th pair of the input and output) and model parameters $\theta$, the uncertainty of the predictive distribution $p(y|x,\mathcal{D})$ can be approximated by considering the data (aleatoric) uncertainty in the likelihood $p(y|\theta,x)$ and model (epistemic) uncertainty in the posterior $p(\theta|\mathcal{D})$.


We first describe the NeRF training from a Bayesian perspective. When training the NeRF, we seek to minimise the total training loss $\mathcal{L}(\mathcal{D};\theta)$ with respect to the NeRF parameters $\theta$ (e.g., using the photometric loss from \eqref{eq:photometric_loss} if only vision is provided). This is equivalent to computing the maximum a-posteriori (MAP) estimate $\hat{\theta}$

\begin{equation}
\begin{aligned}
\hat{\theta} &= \arg\max_{\theta} p(\theta \mid \mathcal{D}) \\
&= \arg\max_{\theta} \big[ \log p(\mathcal{D} \mid \theta) + \log p(\theta) \big]\\
&= \arg\min_{\theta} \big[ -\log p(\mathcal{D} \mid \theta) - \log p(\theta) \big] \\
&= \arg\min_{\theta} \big[  \sum_{n=1}^N \ell(x_n, y_n; \theta) + r(\theta)] \\
&=\arg\min_{\theta} \mathcal{L}(\mathcal{D}; \theta)
\end{aligned}
\label{eq:map}
\end{equation}

where $\ell(x_n, y_n; \theta)=-\log p(y_n | f_\theta(x_n))$ is the loss term that corresponds to the negative log-likelihood for each data sample, and $r(\theta)= - \log p(\theta)$ is the weight regularisation that corresponds to the log-prior.

It can be seen from \eqref{eq:map} that the total training loss can be interpreted as $ \mathcal{L}(\mathcal{D}; \theta)=-\log p(\mathcal{D}|\theta) - \log p(\theta)$. The exponential of the negative training loss then corresponds to the unnormalised posterior $p(\mathcal{D}|\theta)p(\theta)$:
\begin{equation}
p(\theta \mid \mathcal{D}) = \frac{1}{Z} p(\mathcal{D} \mid \theta) p(\theta) 
= \frac{1}{Z} \exp\big(-\mathcal{L}(\mathcal{D}; \theta)\big)
\label{eq:posterior_and_loss}
\end{equation}
where the normalising constant $Z$ is the normalising constant, and is defined as:
\begin{equation}
Z = \int p(\mathcal{D} \mid \theta) p(\theta) \, d\theta
\label{eq:normalise_constant}
\end{equation}

Here, the posterior $p(\theta|\mathcal{D})$ is used for the uncertainty estimation described later in \secref{sec:la}.

\subsection{Laplace Approximation}
\label{sec:la}
Laplace approximation is a technique to estimate the otherwise intractable posterior distribution as a Gaussian function. This allows one to efficiently approximate the posterior distribution, and has been used widely in Bayesian deep learning literature~\cite{papamarkou2024bdl}. In this section, we describe the details of the Laplace Approximation, following the presentation by Daxberger et al.~\cite{daxberger2021laplace}.

First, we take a second-order Taylor Series expansion of the loss function at the MAP estimate $\hat{\theta}$ as follows:

\begin{equation}
	\mathcal{L}(\mathcal{D}; \theta) \approx \mathcal{L}(\mathcal{D}; \hat{\theta}) 
	+ \frac{1}{2} (\theta - \hat{\theta})^\top 
	\mathbf{H}
(\theta - \hat{\theta}),
\label{eq:2ndtaylor}
\end{equation}

where $\mathbf{H}=\nabla_\theta^2 \mathcal{L}(\mathcal{D};\theta) |_{\hat{\theta}}$ is the Hessian matrix at the MAP estimate $\hat{\theta}$. Here, the first-order term $\nabla_\theta \mathcal{L}(\mathcal{D};\theta)|_{\hat{\theta}}^\top(\theta-\hat{\theta})$ does not appear as it is zero at the MAP estimate. 

Substituting the approximation in \eqref{eq:2ndtaylor} into \eqref{eq:normalise_constant}, we obtain:
\begin{equation}
\begin{aligned}
	Z &= \int p(\mathcal{D} \mid \theta) p(\theta) \, d\theta \\
	&= \int \exp(-\mathcal{L}(\mathcal{D}; \theta)) d\theta \\
	&\approx \exp (-\mathcal{L}(\mathcal{D}; \hat{\theta})) \int \exp \left(-\frac{1}{2}(\theta - \hat{\theta})^\top \mathbf{H}(\theta - \hat{\theta})\right)d\theta \\
	&= \exp (-\mathcal{L}(\mathcal{D}; \hat{\theta})) \frac{(2\pi)^{\frac{D}{2}}}{(\det \mathbf{H})^{\frac{1}{2}}}
\end{aligned}
\label{eq:normalise_constant_approx}
\end{equation}

where $D$ denotes the dimensionality of the parameters $\theta$.

We then substitute the Taylor expansion~\eqref{eq:2ndtaylor} and expression of the normalization constant~\eqref{eq:normalise_constant_approx} into the posterior \eqref{eq:posterior_and_loss}:
\begin{equation}
p(\theta | \mathcal{D}) \approx \frac{(\det \mathbf{H})^{\frac{1}{2}}}{(2\pi)^{\frac{D}{2}}} \exp \left(-\frac{1}{2}(\theta - \hat{\theta})^\top \mathbf{H}(\theta - \hat{\theta})\right)
\end{equation}
which corresponds to a Gaussian distribution  $\mathcal{N}(\theta; \hat{\theta}, \mathbf{\Sigma})$ with mean $\hat{\theta}$ and covariance $\mathbf{\Sigma}=\mathbf{H}^{-1}$.

By using Laplace approximation, the uncertainty estimation problem can be formulated as estimating the Hessian matrix $\mathbf{H}$ at the MAP estimate $\hat{\theta}$. We describe how we apply this to the NeRF reconstruction problem in \secref{sec:la_for_nerf}, and techniques to further approximate the Hessian matrix in \secref{sec:hessian_approx}.

\section{Method}

In this section, we present SiLVR, a 3D reconstruction system based on a NeRF representation. An overview of our system is presented in \figref{fig:sys_overview}. SiLVR takes in as input synchronised triplets of a camera image, a lidar depth image and the corresponding surface normals, as well as the sensor trajectory estimated by an online Lidar SLAM system. This trajectory is refined and partitioned into submaps. For each submap, we train a NeRF which combines both vision and lidar data. After training the NeRF, SiLVR estimates the epistemic uncertainty of the NeRF. The final point cloud reconstruction is extracted from the NeRF based on rendered depth, and filtered using the per-point uncertainty estimates. 

We describe our approach to extend the NeRF formulation to include lidar data in \secref{sec:lidar_nerf}, and how to remove the sky reconstruction in \secref{sec:sky_segmentation}. We then introduce the epistemic uncertainty estimation method in \secref{sec:uncertainty}. Finally, we present how SiLVR achieves scalability by partitioning large-scale poses into submaps in \secref{sec:large_scale_pose}.

\subsection{Geometric Constraints from Lidar Measurements}
\label{sec:lidar_nerf}
Image-based 3D reconstruction with NeRF becomes challenging when a surface has a uniform texture and limited multi-view constraints. Lidar measurements are complementary as they can provide accurate depth measurements in such scenarios. In our work, we incorporate the lidar measurements directly into the NeRF optimization. Specifically, we project the lidar point cloud from VILENS-SLAM's pose graph onto the image plane using the camera intrinsics and lidar-camera extrinsics (described in \secref{sec:implementation}) to form a depth image. We denote the lidar depth images as $\mathcal{D}_d$. Each frame of the lidar point cloud is motion-undistorted~\footnote{See implementations at \url{https://github.com/ethz-asl/lidar_undistortion}} to the nearest image's timestamp, and hence synchronised with the image data. Example overlays can be found in \figref{fig:sample_overlay}.

\subsubsection{Lidar Depth Constraints}
\label{sec:method_depth_constraints}
We follow the depth regularisation approach proposed by DS-NeRF~\cite{deng2022depth} for RGB-D cameras. We define the rendering weight distribution $w(t)$ as a discrete probability distribution, with probabilities given by $w_i$, the weights of the ray samples defined in~\eqref{eq:rendering}. The depth regularisation term is obtained by minimising the Kullback-Leibler (KL) divergence between a narrow normal distribution centred at the lidar depth measurement $\mathbf{D}$ and the rendering weight distribution $w(t)$:
\begin{equation}
\mathcal{L}_{\text {Depth-KL}}=\sum_{\mathbf{r} \in \mathcal{D}_d} \mathrm{KL}[\mathcal{N}(\mathbf{D}, \hat{\sigma}) \| w(t)]
\label{eq:depth}
\end{equation}
During training, we apply this regularisation using a coarse-to-fine schedule by progressively reducing the covariance $\hat{\sigma}$ of the target Normal distribution. This encourages the rendering weight distribution $w(t)$ to approach a Dirac delta function. As a result, the density along the ray is encouraged to be concentrated near the lidar depth.






\subsubsection{Surface Normal Constraints from Lidar Measurements}
\label{sec:method_normal_constraints}
\begin{figure}[h]
	\centering
 
	\includegraphics[width=\columnwidth]{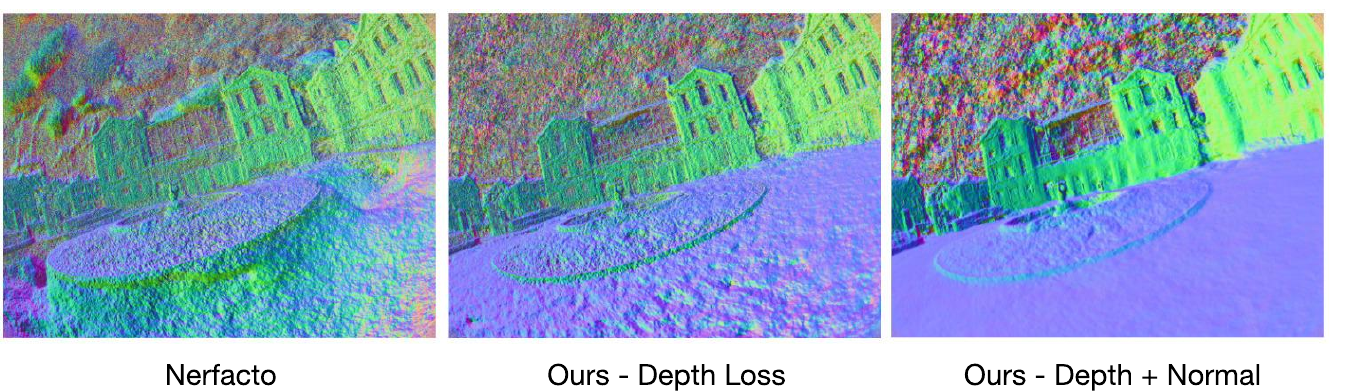}
	\caption{Comparison of surface normal renderings of the Maths Institute. Incorporating normal constraints in addition to depth from lidar improves the smoothness of the reconstruction. Right: The smooth reconstruction of the ground portion highlights this improvement.}    
	\label{fig:surface_normal}
\end{figure}

While the depth loss improves 3D reconstruction, we found that the surface it produces will still contain wavy artefacts in regions where it is expected to be smooth (see \figref{fig:surface_normal}). To mitigate this, we regularise the surface normal of the NeRF with lidar data. For each point $\mathbf{p}$ in the radiance field $R$, its surface normal can be computed as the negative gradient of the volume density $\sigma$. To obtain the training labels for the surface normal, we use the lidar range image and estimate the surface normals by local plane fitting. The surface normals are projected onto the image plane to generate the surface normal images, denoted as $\mathcal{D}_n$, similar to the lidar depth images. Then, we introduce an additional surface normal regularisation loss in the NeRF training, inspired by~\cite{Yu2022MonoSDF}:
\begin{equation}
	 \label{eq:normal}
    \mathcal{L}_{\text {Normal }}=\sum_{\mathbf{r} \in \mathcal{D}_n}\|\hat{N}(\mathbf{r})-\bar{N}(\mathbf{r})\|_1+\left\|1-\hat{N}(\mathbf{r})^{\top} \bar{N}(\mathbf{r})\right\|_1
\end{equation}
Compared to learning-based surface normal estimation from the camera image used in \cite{Yu2022MonoSDF}, our surface normal is estimated from the 3D lidar point cloud and does not suffer from generalisation issues. The effect of the surface normal regularisation can be seen in \figref{fig:surface_normal}.

\subsection{Sky Segmentation}
\label{sec:sky_segmentation}
Since we focus on large-scale reconstructions of outdoor spaces, the sky and clouds are often present in our training image data. Vision-only NeRF reconstruction typically tries to model it as unconstrained floating white or blue points, which become artefacts in the final reconstruction. To remove these ``sky points'' from the training procedure, we used a semantic segmentation network~\footnote{We used Detectron2 from \url{https://github.com/facebookresearch/detectron2}} to obtain a sky mask that is used to regularise the corresponding camera rays to be empty. Specifically, for the rays $\mathbf{r}$ that correspond to the sky mask (denoted as $\mathcal{D}_s$), we minimise the weights of all samples on these rays, similar to~\cite{rematas2022urban}:

\begin{equation}
\mathcal{L}_{\text {Sky}}=\sum_{\mathbf{r} \in \mathcal{D}_s}\sum_{i} w_i^2
\label{eq:sky_loss}
\end{equation}

Combining the loss terms introduced in \eqref{eq:photometric_loss}, \eqref{eq:depth}, \eqref{eq:normal} \eqref{eq:sky_loss}, our overall training loss is

\begin{equation}
	\mathcal{L}=\mathcal{L}_{\text {Colour}} + \mathcal{L}_{\text {Depth-KL}} + \mathcal{L}_{\text {Normal }} + \mathcal{L}_{\text {Sky}}
\end{equation}


\subsection{Epistemic Uncertainty of the NeRF reconstruction}
\label{sec:uncertainty}

We aim to obtain an explicit metric of uncertainty of our NeRF reconstruction. Particularly, following \secref{sec:bayesian_nerf}, we aim to quantify the epistemic uncertainty that is directly related to the covariance of the posterior distribution $p(\theta|\mathcal{D})$ using the Laplace approximation. We first describe the reformulation of the parameters $\theta$, and then we derive the epistemic uncertainty estimate using the approximation.

\subsubsection{Perturbation Field Reformulation}
\label{sec:perturbation}
While a NeRF representation presents convenient advantages for scene compression, its parameters $\theta$ do not directly correspond to the 3D scenes. A change of one parameter in the MLP could change the whole radiance field, and it is difficult to obtain uncertainty for a specific 3D location. This is in contrast to other approaches, such as 3D Gaussian Splatting~\cite{kerbl3Dgaussians}, where the parameters have a direct representation in the world. This introduces additional challenges when estimating the uncertainty of the parameters.
 
 To obtain the spatial uncertainty of a NeRF, we adopt the perturbation field formulation introduced in BayesRays~\cite{goli2023bayesrays}. Specifically, we construct the perturbation field $\mathcal{P}$ of every 3D coordinate $\mathbf{x}$. We define the perturbation field as $\mathcal{P}_{\theta_P}(\mathbf{x}):\mathbb{R}^3 \rightarrow \mathbb{R}^3$, where $\theta_P$ denotes the parameters in the perturbation field. A 3D coordinate's perturbation can be obtained using trilinear interpolation:
\begin{equation}
    \mathcal{P}_{\theta_P}(\mathbf{x}) = \text{Trilinear}(\mathbf{x}, \theta_P)
\end{equation}

When estimating the uncertainty, we modify the volume rendering procedure by adding the perturbation field to the point samples $\mathbf{p}_i$ along the ray $\mathbf{r}$ to produce the new point sample $\mathbf{p}_i'=\mathbf{p}_i + \mathcal{P}_{\theta_P}(\mathbf{p}_i)$. The new point colour and volume density $(\mathbf{c}_i',\sigma_i')=R(\mathbf{p}',\mathbf{d})$, and the new pixel colour $\hat{\mathbf{C}'}(\mathbf{r})$ can then be computed using \eqref{eq:pixel_colour} and \eqref{eq:rendering}.

The introduction of the perturbation field enables us to obtain uncertainties of a specific location in the 3D space, which is crucial for our application. From the Bayesian formulation of our problem, this means that the parameter $\theta$ in the posterior $p(\theta|\mathcal{D})$ (whose covariance we estimate as our model uncertainty) is not the NeRF parameters $\theta_N$ (MLP weights and the hash grids), but the perturbation field $\theta_P$ (perturbation value stored in the grid vertices.

\subsubsection{Epistemic Uncertainty Estimation using Laplace Approximation}
\label{sec:la_for_nerf}
We estimate the epistemic uncertainty of the NeRF reconstruction by estimating the covariance of the posterior $p(\theta|\mathcal{D})$. Using the Laplace approximation technique described in \secref{sec:la}, we estimate the otherwise intractable posterior distribution as a Gaussian function, and then use its covariance as the estimated uncertainty of our reconstruction. 

Specifically, the mean of the Gaussian $\hat{\theta}$ is the MAP estimate of the parameters $\theta$. Here, the parameters $\theta$ that we are estimating are the perturbation field $\theta_P$ introduced in \secref{sec:perturbation}. Assuming that the NeRF reconstruction has converged to a local minima after training, a small perturbation should not cause the reconstruction to deteriorate, and hence the MAP estimate of the perturbation field is $\mathbf{0}$ (no perturbation). Then, the major computation needed is to determine the covariance, or the inverse of the Hessian. If we assume the prior is a zero-mean Gaussian $p(\theta)=\mathcal{N}(\theta;\mathbf{0},\gamma^2 \mathbf{I})$, then the full Hessian at the location of the MAP estimate is
\begin{equation}
	\begin{aligned}
		\mathbf{H} &=\nabla_\theta^2 \mathcal{L}(\mathcal{D};\theta) |_{\hat{\theta}} \\
		&= -\gamma^{-2} \mathbf{I} - \sum_{n=1}^N \nabla_\theta^2 \log p(y_n | f_\theta(x_n))|_{\hat{\theta}}\\
	\end{aligned}
	\label{eq:hessian}
\end{equation}

\subsubsection{Approximation of the Hessian Matrix}
\label{sec:hessian_approx}

While the second term in the Hessian from \eqref{eq:hessian} is usually intractable, it can be approximated by the Fisher information matrix~\cite{amari1998natural}:
\begin{equation}
	\begin{aligned}
		\mathbf{H} \approx -\gamma^{-2} \mathbf{I} - \sum_{n=1}^N \mathbf{J}\mathbf{J}^\top
	\end{aligned}
\end{equation}
%
where $\mathbf{J}=\nabla_\theta \log p(y_n | f_\theta(x_n))|_{\hat{\theta}} = -\nabla_\theta \ell(x_n, y_n; \theta)$ is the Jacobian matrix of the NeRF model.


Since the Fisher information matrix is still expensive to compute, we can make a further assumption that each parameter (perturbation field) is independent of the others, and use a diagonal approximation to the Hessian:
\begin{equation}
	\begin{aligned}
		\mathbf{H} \approx  -\gamma^{-2} \mathbf{I} - \text{diag}(\mathbf{J}\mathbf{J}^\top)\\
	\end{aligned}
\end{equation}

This means that for the Hessian matrix, its diagonal elements $H_{ii}$ can be computed as 
\begin{equation}
	\label{eq:hessian_final}
	H_{ii} \approx \sum_{j=1}^n \left(\frac{\partial \ell_j}{\partial \theta_i}\right)^2 + \gamma^{-2}
\end{equation}

The Hessian matrix is initialised as all zeroes; i.e., infinite covariance. After accumulating the gradients, the prior term $\gamma^{-2}$ in \eqref{eq:hessian_final} ensures that the final Hessian matrix is positive definite and the covariances are bounded. As a result, the parameters that are not involved in the outputs (the rendered pixels or depth) will have very high epistemic uncertainty because changing them will not change the outputs and the training loss. Since the outputs are rendered according to the rays from the training images, the unobserved regions can be identified as having very high uncertainty and can be filtered out. Identifying unobserved areas (similar to the unknown space in occupancy mapping) is particularly important for NeRF, as the underlying MLP can output arbitrary colour and density in some locations, leading to ``hallucinated'' reconstruction points.

\subsubsection{Uncertainty for Heterogeneous Sensors}
The NeRF model is trained with a total training loss function $\mathcal{L}(\mathcal{D;\hat{\theta}})$ that contains the photometric loss $\mathcal{L}_{\text {Colour}}$ and the depth loss $\mathcal{L}_{\text{Depth-KL}}$. This means that the Jacobian $\mathbf{J}$ can be decoupled into a colour component $\mathbf{J}_{\text{Colour}}$ and depth component $\mathbf{J}_{\text{Depth-KL}}$, from which we can then approximate the Hessian that corresponds only to the visual information  $\mathbf{H}_{\text{Colour}}$, and the Hessian that corresponds to only to the lidar depth information $\mathbf{H}_{\text{Depth-KL}}$.
Therefore, we can compute the epistemic uncertainty for \textit{each observation modality} by changing the loss function. Note that the total training loss function contains other terms including the surface normal loss $\mathcal{L}_{\text {Normal}}$. In our study, we focus on just the photometric loss $\mathcal{L}_{\text {Colour}}$ and depth loss $\mathcal{L}_{\text{Depth-KL}}$. This is because these two losses are the dominant components of the total loss $\mathcal{L}(\mathcal{D;\hat{\theta}})$ (with our weighting coefficients for each loss chosen).

The nature of the different sensor modalities leads to different uncertainty characteristics. The visual uncertainty captures areas that can be geometrically perturbed while not changing the colour. As later shown in \figref{fig:unc-ablation-plots}, ``low'' visual uncertainty corresponds to distinct visual features and high-frequency areas. ``High'' visual uncertainty typically corresponds to areas with uniform texture, since perturbing a point in an area with similar colours can lead to small changes in the rendered colour. Here, the characteristics of the visual uncertainties are similar to those in SfM where visual features are used as landmarks for Bundle Adjustment, whereas uniform texture areas are often not mapped.

In comparison, low lidar depth uncertainty often appears in regions with abundant lidar observation---whether there are visual features or not. This means a road surface with uniform texture can have \textit{lower} lidar depth uncertainty but \textit{higher} visual uncertainty. Low lidar depth uncertainty is also observed at
object boundaries, since perturbing that point can lead to a drastic change in the depth (from foreground depth to background depth). Regions with high lidar depth uncertainty are often areas where there are fewer lidar observations, such as the sky, distant regions that are beyond the lidar's sensing range, and regions outside the lidar's FOV.

The decoupling of the uncertainties enables a principled interpretation of the lidar-visual reconstruction. We can identify parts of the reconstructions that are reliable (surface with visual features, and/or abundant lidar observations) and unreliable (no lidar observation and uniform-textured surfaces), given each sensor modality's own characteristics.






\subsection{Large-scale Pose Trajectory Estimation}
\label{sec:large_scale_pose}

\subsubsection{Refining Lidar SLAM trajectory with Bundle Adjustment}
\label{sec:pose_refinement}
Providing the NeRF reconstruction method with accurate camera poses is crucial, as their accuracy directly impacts the fidelity of the reconstructed model. A popular approach used in most NeRF works is to estimate camera poses using (offline) Structure-from-Motion methods such as COLMAP~\cite{schoenberger2016colmap}. However, we observed the following limitations when testing COLMAP: (1) long computation times,  especially for large image collections collected spanning a long trajectory (e.g., 3000 images can take more than 1 hour (as shown in \tabref{tab:pose_ablation}), and (2) inability to register all frames into one global map when there is limited visual overlap between the images. These issues undermine the goal of building complete, large-scale, globally consistent maps.

In our work, we use our lidar-inertial odometry and SLAM system VILENS~\cite{wisth2023vilens}. While VILENS achieves state-of-the-art results for lidar-based online motion tracking, we found that the camera poses obtained are less precise than those of COLMAP. This results in \textit{blurring} artefacts in the images rendered by the NeRF model. Several works~\cite{tancik2022block,azinovic2022neural} use noisy pose inputs and then jointly refine the poses within the NeRF optimization to generate better results. However, as shown later in \tabref{tab:pose_ablation}, our experiments showed that this pose-refinement approach can produce results which are inferior to using poses estimated by COLMAP.


To overcome these limitations, we propose to use the pose trajectory from VILENS SLAM as a prior and refine it by running Bundle Adjustment using COLMAP~\cite{schoenberger2016colmap}. Specifically, we first run the feature extraction and matching on the dataset, and then use the Lidar SLAM poses to triangulate visual feature points, and run a few iterations of BA.
This method is faster than a typical incremental SfM, as it reduces the incremental Structure-from-Motion to a Bundle Adjustment problem. More importantly, having an accurate prior for all the image poses means that COLMAP will be able to produce a full solution and does not fail to register some of the images---as would be the case for SfM without initialisation. 
For a mission spanning over 20 minutes, our COLMAP-with-prior pipeline achieves similar rendering quality, while typically taking half the computation time of a standard COLMAP run. The computation time is similar to the time required by a robot to collect the data, making it more suitable for robotic applications. 

After COLMAP computation, we rescale the trajectory using $\mathit{Sim}(3)$ Umeyama alignment to the Lidar-SLAM trajectory, so that the final trajectory is metrically scaled. This step is crucial because the lidar measurements used in \secref{sec:method_depth_constraints} are also metric. The depth regularisation cannot be used if the scale of the scene and the scale of the depth are not consistent. 

\subsubsection{Submapping of Pose Trajectory}
To divide the whole map into smaller manageable areas, we partition the entire trajectory into shorter trajectories, which we define as submaps.
The submaps are clustered considering image visibility rather than using space partitioning or distance-based clustering~\cite{tao2024silvr}. The goal is to exclude an image from a submap if it does not contribute to the submap reconstruction---for example, if the scene observed is not visible from the other images in the submap. 

We formulate the clustering problem as a graph partitioning problem, where each node is an image, and the edges between the nodes are weighted by a similarity score. We measure the similarity between two images using a co-visibility metric. Two images are co-visible if some feature points on one image can be viewed from the other image. Specifically, the co-visibility metric for an image pair is computed as the number of visual feature points computed by COLMAP that appear in both images.
After constructing the graph, we use the Normalised Cuts algorithm~\cite{shi2000normalized} to obtain a partitioning which minimally breaks edges; i.e., to remove the link between unrelated nodes. In practice, this means that images that are co-visible are grouped together, and image pairs that have less visual overlap are identified as the submap boundary. 



Once we divide the full map into submaps as a set of clustered images, we independently train each NeRF submap. After training, we compute the epistemic uncertainty of the reconstruction. We can export a point cloud by rendering colour and depth using the training data rays, and we filter out points that have high uncertainty. Rendering at the submap intersections can be obtained by combining two submap renderings in the image space where the weights are determined based on the distances to the neighbouring submap, following Block-NeRF~\cite{tancik2022block}.



\begin{figure}[h]
	\centering
	\includegraphics[width=0.99\columnwidth]{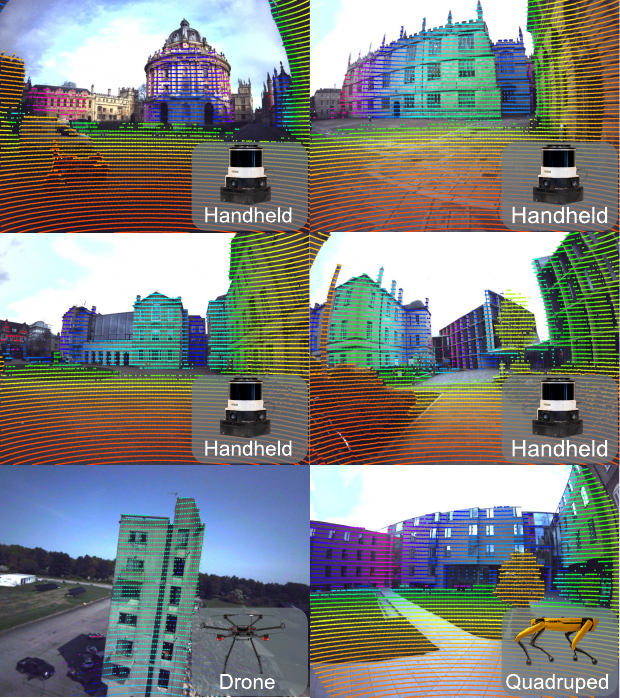}
	\caption{Sample Data from our diverse robotic datasets. Here, each image is overlaid with a projected lidar point cloud to demonstrate the accuracy of the sensor calibration.}    
	\label{fig:sample_overlay}
\end{figure}

\begin{table}[h]

    \centering
    \begin{tabular}{c c c c}
    \toprule
        Site Name & Robotic Platform & GT   \\
    \midrule
        HB Allen Centre & BD Spot & Leica BLK360  \\
        Fire Service College & DJI M600 Drone  & Leica BLK360 \\

        Radcliffe Observatory Quarter & Handheld Frontier  & Leica RTC360 \\
        Bodleian Library & Handheld Frontier  & Leica RTC360 \\
    \bottomrule
    \end{tabular}
    \caption{Details of real-world datasets that are used for evaluation.} 
    \label{tab:datasets}
\end{table}

\section{Experimental Setup}

\subsection{Hardware and Datasets}
We evaluate our methods using a custom perception unit called Frontier shown in \figref{fig:sys_overview}. It includes three 1.6 MP fisheye colour\footnote{To produce RGB images, we debayer and white-balance the raw bayered images using \url{https://github.com/leggedrobotics/raw_image_pipeline}} Alphasense cameras (from Sevensense Robotics AG) on 3 sides of the device, as well as a synchronised IMU. The 3-camera setup enables omnidirectional NeRF mapping from a single walking pass through a test site. We installed a Hesai QT64 lidar (104\degree~FOV, 60 metres maximum range) on top of the device. We deployed the Frontier in different modes---onboard a legged robot (Boston Dynamics Spot), a drone (DJI M600, with the system described in \cite{border2024osprey}), or simply handheld (using the Oxford Spires dataset~\cite{tao2025spires}. Some data was collected with the Frontier mounted on a human operator's backpack).

In addition, we used a survey-grade terrestrial lidar
scanner (TLS) to obtain a millimetre-accurate point cloud which we later used to create a reference ground truth model. We use either the entry-level Leica BLK360 or the professional-grade Leica RTC360 to obtain ground truth maps of the sites.

We tested our method using data collected in the following sites: H B Allen Centre (HBAC), Fire Service College (FSC), Radcliffe Observatory Quarter (ROQ), and the Bodleian Library, all in Oxford, UK. The large-scale sites, ROQ and the Bodleian Library, cover areas of \( \text{5,000}~\text{m}^2 \) and \( \text{15,000}~\text{m}^2 \), respectively. The hardware details are in \tabref{tab:datasets}, and some sample lidar-camera overlays are shown in \figref{fig:sample_overlay}.


\begin{figure*}[t]
	\centering
	\includegraphics[width=2\columnwidth]{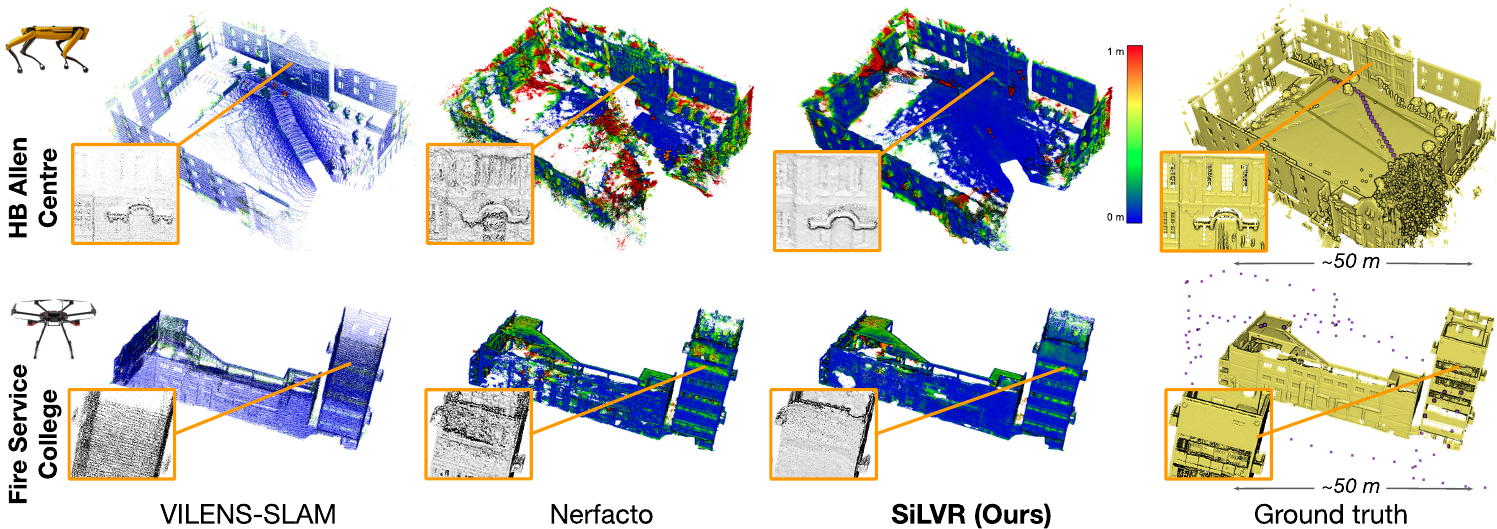}
	\caption{Comparison of reconstruction quality of VILENS-SLAM, Nerfacto (vision-only) and our approach in small-scale scenes. Reconstructions are coloured using the point-to-point distance between the respective reconstructions and the ground truth scan with increasing error from blue (0m) to red (1m). The trajectory is shown in purple and overlaid on the ground truth scan captured using a Leica BLK360. The zoomed-in views show the difference in reconstruction quality. Overall, our approach is more complete w.r.t lidar-only reconstruction, and geometrically more consistent w.r.t vision-only reconstruction.}
	\label{fig:reconstruction_eval_small_scale}
\end{figure*}

\begin{table}[h]
	\caption{\small{ Evaluation of 3D Reconstruction Quality of Small Scenes}}
   	\setlength{\tabcolsep}{3pt} 
	\centering
	\begin{tabular}{ l c c c c c}
		\toprule
		Method& Accuracy$\downarrow$ & Completeness$\downarrow$ &  \multicolumn{2}{c}{PSNR$\uparrow$} & SSIM$\uparrow$ 
		\\
            &(m)&(m)&train&test&test
            \\
            \hline
		\addlinespace
		\textbf{Oxford HBAC}
		\\
		\hline
		\addlinespace
		VILENS-SLAM &\textbf{0.05}& 0.25& /&/&/
		\\
		Nerfacto mono&0.49&5.40 &\textbf{32.6}&19.5&0.65
            \\
            Nerfacto 3-cam&0.28&0.40&29.8&20.6&\textbf{0.74}
		\\
		Ours mono& 0.30&4.60&31.0&\textbf{21.2}&\textbf{0.74}
            \\
            Ours 3-cam &0.09&\textbf{0.18}&28.8&19.7&\textbf{0.74}
		\\
  		\hline
		\addlinespace
            \textbf{FSC} 
		\\
		\hline
		\addlinespace
		VILENS-SLAM		&\textbf{0.08} &\textbf{0.08}&/&/&/
            \\
		Nerfacto mono &0.14&0.11 &\textbf{28.8}&\textbf{19.1}&\textbf{0.76}
		\\
		Ours mono&0.11 &0.09&27.7&\textbf{19.1}&0.75
		\\
		\bottomrule
		\addlinespace
	\end{tabular}
	\label{tab:reconstruction_eval_small_scale}
\end{table}


\subsection{Implementation Details}
\label{sec:implementation}
\subsubsection{Data Collection and Processing}
When collecting the data, we use VILENS~\cite{wisth2023vilens}, a lidar-inertial SLAM system running online to estimate a globally consistent trajectory and to motion correct the lidar measurements. The SLAM trajectory estimated online can also be further optimised using Bundle Adjustment as described in \secref{sec:pose_refinement}. This improves the visual reconstruction quality, as shown later in \tabref{tab:pose_ablation}. Individual lidar scans are projected to form a sparse depth image coinciding with the camera image (i.e., the same camera pose and intrinsic parameters). Surface normals of the lidar points are estimated as the surface normal of the polygon formed by the current point's neighbouring points in the lidar range image using Newell's method, and projected as a sparse normal image. We use the calibrations provided by the Oxford Spires dataset~\cite{tao2025spires}. In this dataset, the intrinsics and extrinsics of the set of cameras are estimated using Kalibr~\cite{furgale2013unified}, and a single extrinsic transformation between the three cameras and the lidar is estimated using DiffCal~\cite{fu2023extrinsics}. When running COLMAP, we further optimise the camera intrinsics produced by Kalibr.

\subsubsection{NeRF Reconstruction}
Our NeRF reconstruction system extends Nerfacto, which is a specific vision-only pipeline implemented within the Nerfstudio framework~\cite{nerfstudio}. Nerfacto's rendering quality is comparable to state-of-the-art methods such as MipNeRF-360~\cite{barron2022mipnerf360} while achieving a substantial acceleration in reconstruction speed as it also incorporates efficient hash encoding which was proposed by the authors of Instant-NGP~\cite{mueller2022instant}. The scene contraction, proposed in~\cite{barron2022mipnerf360}, is also used to improve memory efficiency and to represent scenes with high-resolution content near the input camera locations. The contraction function non-linearly maps any point in space into a cube of side length 2, and represents the scene within this contracted space. In real-world outdoor environments, there are often large variations in exposure and lighting conditions. Because of this, we use a per-frame appearance encoding for each image, similar to~\cite{rematas2022urban,martinbrualla2020nerfw}. To train the NeRF model, we used an Nvidia RTX 4090. One training iteration takes 4096 rays.

\begin{figure*}[t]
	\centering
	\includegraphics[width=2.0\columnwidth]{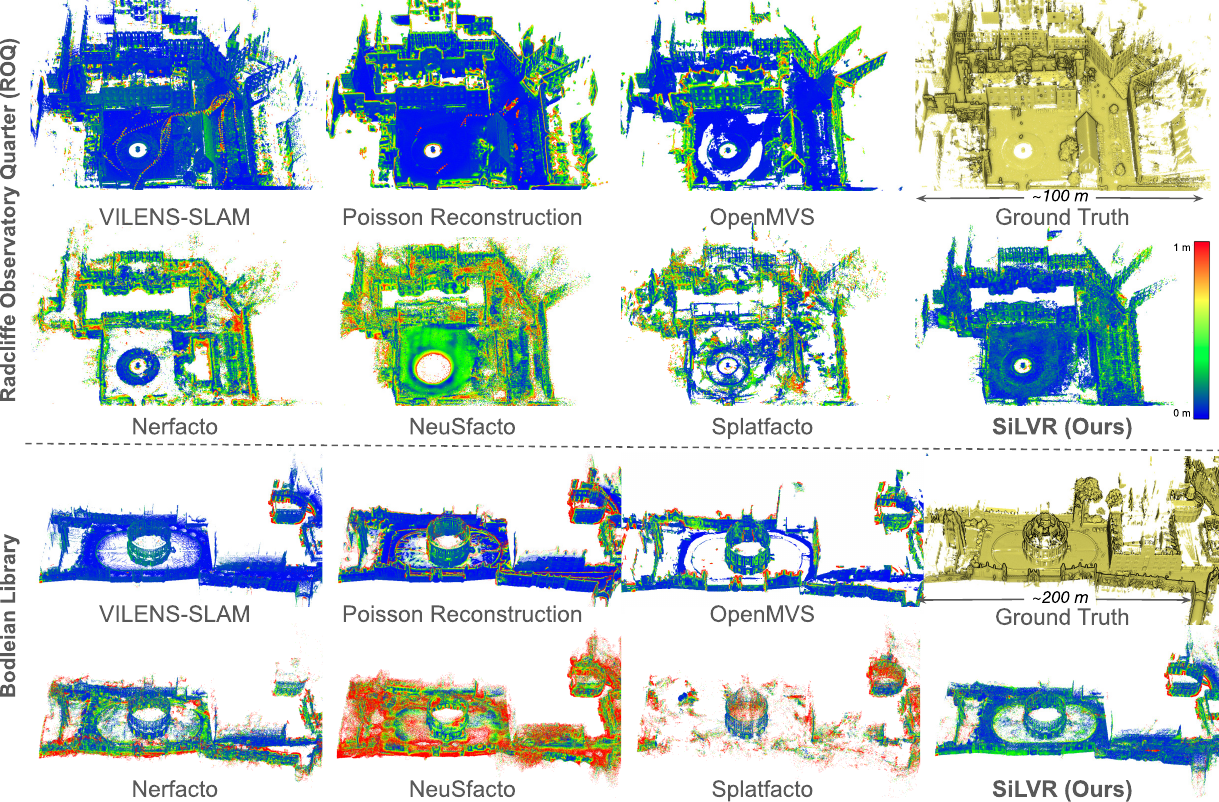}
	\caption{Comparison of reconstruction quality of our method and other baseline methods in two real-world large-scale scenes. Among the radiance field baseline methods, SiLVR's reconstruction is the most accurate and complete, especially on the ground where there are insufficient visual constraints.}
	\label{fig:reconstruction_eval_large_scale}
\end{figure*}

\subsection{Evaluation Details}
\subsubsection{3D Reconstruction Metrics}
To evaluate the geometry of the reconstruction, we report \textit{Accuracy} and \textit{Completeness} following the conventions of the DTU dataset~\cite{aanaes2016large_dtu}. Accuracy is measured as the point-to-point distance from the reconstruction to the (ground truth)  reference 3D model and indicates the reconstruction quality. Completeness is the distance from the point-wise reference to the reconstruction and shows how much of the surface has been captured by the reconstruction. 

In addition, we also compute \textit{Precision} and \textit{Recall} with a pre-defined error threshold. A point in the reconstruction which is below this threshold can be considered to be a true positive. We use both 5cm and 10cm for the threshold following \cite{tao2025spires}.



\begin{figure}[h]
	\centering
	
	\includegraphics[width=1\columnwidth]{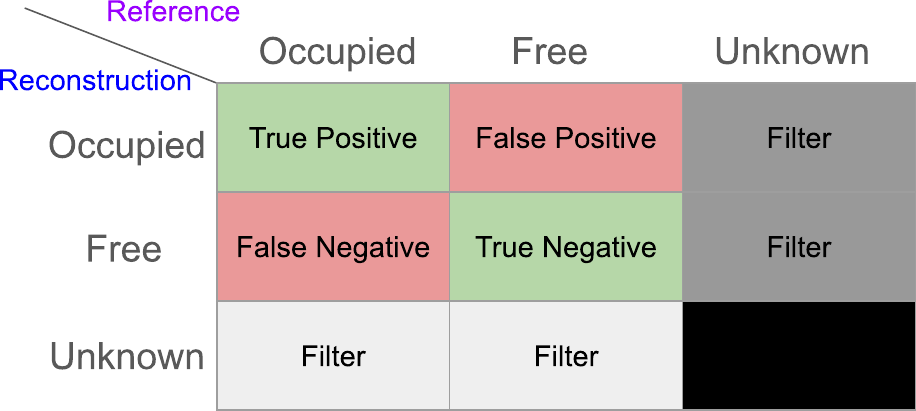}
	\caption{Classification of different occupancy categories for the reconstruction and reference models.}    
	\label{fig:occ_analysis}
\end{figure}

\subsubsection{Map Filtering for Fair Evaluation}
\label{sec:map_filtering}
Perfect accuracy and completeness scores (of zero in both cases) would be achieved if the two point clouds were identical, and any deviation is penalised by a higher value. In practice, the ground truth model and the reconstruction do not perfectly overlap, as they scan slightly different parts of the scene from different viewpoints. Two situations can occur which do not correspond to mapping error:
\begin{enumerate}
    \item Missing regions in the ground truth map: the ground truth map can have undetected areas of the scene that were captured in the Frontier data sequence. This would lead to an artificially higher accuracy score for the Frontier data in such regions. These are in effect \textit{false positives} as shown in \figref{fig:occ_analysis}.
    \item Extra regions in the ground truth map: the ground truth map can contain areas that the Frontier device did not scan. In this case, the NeRF reconstruction of these extra regions will be missing. These are undesirable \textit{false negatives} as shown in \figref{fig:occ_analysis}, and the completeness score would again be artificially higher than it should be.
\end{enumerate}

These overestimated error measures are typically much higher than the errors which occur in well-defined regions (both for the TLS ground truth and the Frontier data), and can then skew the results metrics. This makes comparison between different reconstruction methods difficult.

To address this issue, we filter the non-overlapping regions that we consider should not be included in the evaluation --- for both the reconstruction and the ground truth. Specifically, our evaluation system first builds an occupancy map of the ground truth reconstruction using Octomap~\cite{hornung13octomap}. We then remove points in the reconstruction that are not in the octree (i.e., in the \textit{unknown} space). Similarly, we build an occupancy map of the lidar point clouds, and remove ground truth points within the unknown space. Manual filtering is also applied for regions that are inside the buildings.


\subsubsection{Rendering Metrics}
We evaluate the visual quality of the reconstructions by reporting the Peak Signal-to-Noise Ratio (PSNR) and Structural Similarity Index (SSIM)~\cite{wang2004image}, which are standard metrics in the radiance field literature. Note that our images have variable exposure times which lowers the test PSNR even if the reconstructed images have a very high degree of photorealism. 


\begin{table*}[t]
\caption{ Evaluation of 3D reconstruction quality. Classical methods (Lidar SLAM, Poisson Reconstruction and MVS) and radiance file methods are grouped separately. The best results in each group are indicated in bold.}
\setlength{\tabcolsep}{4pt} 
\centering

\begin{tabular}{l c c c c c c c c c c c c}
\toprule
Method & Accuracy$\downarrow$ & Completeness$\downarrow$ & \multicolumn{3}{c}{5cm} & \multicolumn{3}{c}{10cm} & \multicolumn{2}{c}{PSNR$\uparrow$} & SSIM$\uparrow$ & LPIPS$\downarrow$\\
& (m) & (m) & Precision & Recall & F-Score & Precision & Recall & F-Score & train & test & test\\
\midrule
\multicolumn{13}{l}{Radcliffe Observatory Quarter (ROQ)}\\
\midrule
VILENS-SLAM &\textbf{0.077} & \textbf{1.214} & \textbf{0.552} & \textbf{0.367} & \textbf{0.441} & \textbf{0.832} & \textbf{0.625} & \textbf{0.714} & / & / & /& /\\
Poisson Reconstruction & 0.146 & 1.768 & 0.406 & 0.274 & 0.327 & 0.658 & 0.558 & 0.604 & / & / & /& /\\
OpenMVS & 0.123 & 1.570 & 0.460 & 0.353 & 0.399 & 0.688 & 0.495 & 0.576 & / & / & /& /\\
\addlinespace
Nerfacto 3-cam & 0.916 & 2.272 & 0.220 & 0.072 & 0.109 & 0.392 & 0.189 & 0.256 &\textbf{25.71}&	\textbf{20.92}&	\textbf{0.714}	&\textbf{0.490}\\
Splatfacto 3-cam & 0.478 & 2.415 & 0.240 & 0.044 & 0.074 & 0.395 & 0.151 & 0.218 & 21.96&	19.95&	0.712	&0.514 \\
NeuSfacto 3-cam & 0.699 & 2.763 & 0.051 & 0.021 & 0.030 & 0.115 & 0.098 & 0.106 & 21.17&	16.97	&0.521&	0.549 \\
\textbf{SiLVR (Ours)} & \textbf{0.095} & \textbf{1.803} & \textbf{0.416} & \textbf{0.150} & \textbf{0.221} & \textbf{0.699} & \textbf{0.344} & \textbf{0.461} &24.73	&20.90	&0.653	&0.551\\
\midrule

\multicolumn{13}{l}{Bodleian Library}\\
\midrule
VILENS-SLAM & 1.017 & \textbf{0.736} & \textbf{0.324} & 0.098 & 0.150 & \textbf{0.518} & 0.290 & \textbf{0.372} & / & / & /& /\\
Poisson Reconstruction & 1.256 & 1.230 & 0.239 & 0.104 & 0.145 & 0.400 & \textbf{0.334} & 0.364 & / & / & /& /\\
OpenMVS & \textbf{0.955} & 2.257 & 0.223 & \textbf{0.129} & \textbf{0.163} & 0.429 & 0.280 & 0.339 & / & / & /& /\\
\addlinespace
Nerfacto 3-cam & 2.841 & 1.124 & 0.092 & 0.030 & 0.045 & 0.190 & 0.132 & 0.156 & \textbf{28.92}	&\textbf{23.03}&	0.827	&\textbf{0.715} \\
Splatfacto 3-cam & 13.532 & 1.275 & 0.020 & 0.004 & 0.007 & 0.044 & 0.027 & 0.033 & 23.92&	22.19&	\textbf{0.850}	&0.748\\
NeuSfacto 3-cam & 2.656 & \textbf{1.074} & 0.015 & 0.007 & 0.010 & 0.035 & 0.042 & 0.038 &24.00&	20.61&	0.619	&0.731\\
\textbf{SiLVR (Ours)} & \textbf{1.292} & 1.532 & \textbf{0.129} & \textbf{0.041} & \textbf{0.063} & \textbf{0.276} & \textbf{0.170} & \textbf{0.211} & 28.00&22.94	&0.754	&0.779\\
\bottomrule
\addlinespace
\end{tabular}
\label{tab:reconstruction_eval_large_scale}
\end{table*}


\section{Experimental Results}

\subsection{Evaluation of the 3D Reconstruction}

We perform a quantitative evaluation of our method using real-world datasets captured by different robotic platforms. Of the evaluation datasets used, HBAC and FSC are small-scale scenes (just one building or a single enclosed space), while ROQ and Bodleian Library are large-scale scenes (connected building complexes). We compare the output point cloud reconstructions from the following algorithms:
\begin{enumerate}
    \item VILENS-SLAM: lidar point clouds are registered with poses computed by an odometry system VILENS~\cite{wisth2023vilens} and pose graph optimisation~\cite{ramezani2020slam}
    \item Poisson Reconstruction~\cite{kazhdan2006poisson}: surface reconstruction using point clouds from VILENS-SLAM 
    \item OpenMVS~\footnote{Available at \url{https://github.com/cdcseacave/openMVS}}: multi-view stereo reconstruction  
    \item Nerfacto~\cite{nerfstudio}: vision-only radiance field reconstruction using volume density
    \item NeuSfacto~\cite{wang2021neus}: vision-only radiance field reconstruction using SDF
    \item Splatafacto~\cite{ye2023splatfacto}: vision-only radiance field reconstruction using 3D Gaussians which are initialised using SfM visual features from COLMAP
    \item SiLVR: Our proposed method using photometric loss, depth loss, and surface normal loss
\end{enumerate}

Note that all the methods (except VILENS-SLAM and Poisson reconstruction) use the same set of poses and input images. For the large-scale datasets ROQ and Bodleian Library, we use the same submap partitioning for all the radiance field approaches (Nerfacto, NeuSfacto, Splatfacto, and SiLVR).

We summarise the quantitative results in \tabref{tab:reconstruction_eval_small_scale} and \tabref{tab:reconstruction_eval_large_scale}, and show the 3D reconstructions in \figref{fig:reconstruction_eval_small_scale} and \figref{fig:reconstruction_eval_large_scale}. Among all methods, the lidar-only method VILENS-SLAM is the most accurate and complete. 
OpenMVS produces much more accurate and complete reconstructions compared to the radiance field reconstruction, but produces a poor reconstruction of the ground compared to lidar-based methods. This is expected as there is little texture on the ground.
All the radiance field reconstructions are less accurate and less complete compared to VILENS-SLAM and OpenMVS. Nerfacto fails to estimate most of the ground geometry accurately in ROQ (the reconstruction is below the ground and is filtered by the occupancy map described in \secref{sec:map_filtering}). Compared to Nerfacto, NeuSfacto reconstructs the ground surface better, but still at an incorrect height compared to the ground truth. This shows that while the SDF formulation poses a geometric prior on the scene (enforcing that there should be a surface rather than an arbitrary volumetric field), it still cannot estimate the surface accurately if there are insufficient visual constraints. The 3D Gaussians exported by Splatfacto also cannot reconstruct the ground accurately. These 3D Gaussians are located mostly on the visual features of the sites---since they are initialised by the COLMAP feature points. 

\begin{figure*}[t]
	\centering
	\includegraphics[width=1.98\columnwidth]{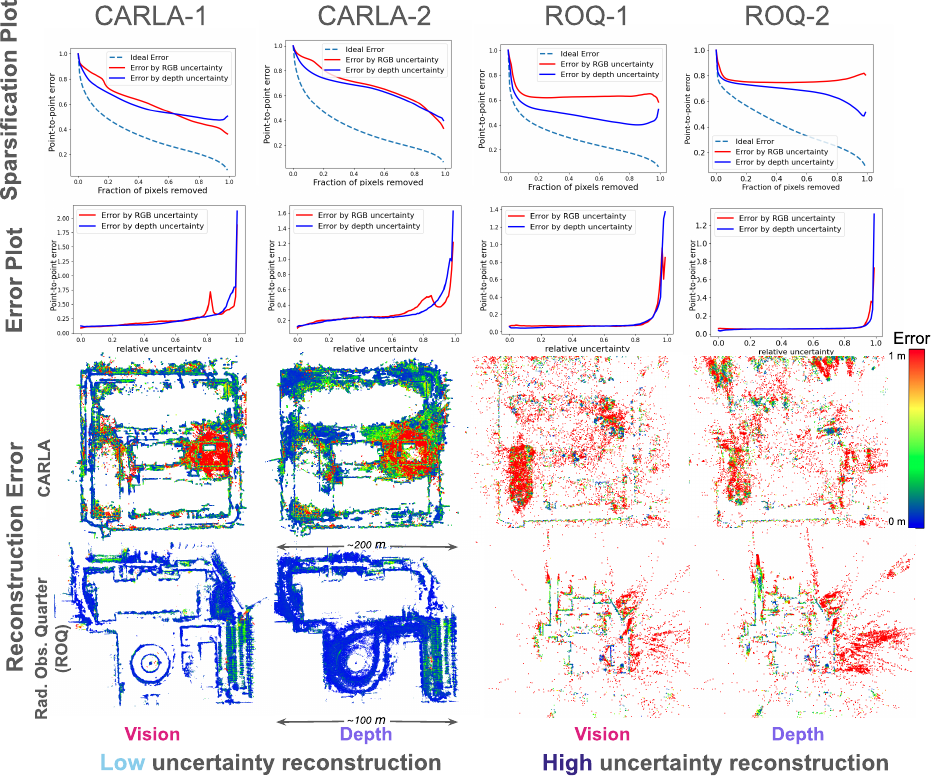}
	\caption{Qualitative and quantitative evaluation of epistemic uncertainty estimates using both synthetic (CARLA) and real-world (ROQ) datasets. The Sparsification Plot shows how the (normalised) reconstruction error decreases as more uncertain points are removed, and how close it is to the oracle error reduction curve. The Error Plot additionally shows the distribution of error at different uncertainty estimates. Low visual uncertainty corresponds to visual features which are well constrained by the image view constraints, similar to the SfM and Visual SLAM systems. Low lidar depth uncertainty indicates that there is abundant lidar observation, and hence the geometry is also constrained.}
\label{fig:unc-ablation-plots}
\end{figure*}

Compared to the vision-only methods, SiLVR incorporates lidar measurements and has significantly better reconstruction fidelity, especially on the ground. Compared to VILENS-SLAM, SiLVR achieves more complete reconstruction for the shorter sequences (e.g., HBAC in \figref{fig:reconstruction_eval_small_scale}) since it uses dense visual information. When there are many accumulated lidar points (which is the case for the large-scale datasets in \figref{fig:reconstruction_eval_large_scale}), this advantage is less prominent.

Regarding the rendering quality, Nerfacto achieves the best results among the radiance field reconstructions. We found that NeuSfacto's training takes longer than all the other methods, and the rendering quality is worse than the other methods. SiLVR achieves a balance between the rendering quality and the 3D reconstruction quality.



\subsection{Evaluation of Epistemic Uncertainty Estimation}
We evaluate the epistemic uncertainty estimates using both synthetic data and real-world data. The synthetic data is generated using the CARLA simulator~\cite{Dosovitskiy17carla}, which provides perfect pose trajectories and ground truth maps. We simulated a vehicle with a lidar and three cameras in a configuration similar to the Frontier. Meanwhile, the real-world dataset used in this section is the Radcliffe Observatory Quarter (ROQ).
\subsubsection{Evaluation Metrics}
We evaluate the epistemic uncertainty estimates using the Sparsification Plot which has been used for evaluating confidence estimates in the literature~\cite{ilg2018uncertainty,Eldesokey2020pncnn,goli2023bayesrays}. The Sparsification Plot is used to evaluate how the uncertainty estimates coincide with the actual errors (in our case, point-to-point distance to the ground truth). In these plots, reconstruction points with the highest uncertainty are gradually removed, and the average errors of the remaining reconstruction points are calculated to form a graph. If the uncertainty estimates align perfectly with the actual errors, then the reconstruction points with the highest errors are always removed first, which leads to the steepest decrease of the remaining error as the most uncertain points are being removed. This ideal sparsification curve is referred to as \textit{Oracle Sparsification}. In practice, the uncertainty estimates do not align with the actual errors perfectly. Specifically, when a reconstruction point has a higher uncertainty but lower error compared to another point, the uncertainty estimates are considered not perfect. The area between the sparsification and its oracle indicates how different the uncertainty estimates are from the ideal ones, and can then be used to compute the Area Under Sparsification Error (AUSE). A smaller difference between the sparsification and its oracle results in a lower AUSE, and indicates that the uncertainty estimates are better because they align better with the actual errors.

In addition to the Sparsification Plot which focuses on the error of the \textit{remaining} reconstructions, we also analyse the error of the reconstruction that is \textit{being removed} (according to the uncertainty). This is achieved by plotting the errors of the reconstruction that have different levels of uncertainties, which we refer to as the Error Plot. While the errors in the Sparsification Plots are normalised (since AUSE is scale-invariant), we use metric errors in the Error Plot to keep the scale information.    




\subsubsection{Results}
In \figref{fig:unc-ablation-plots}, we evaluate the decoupled epistemic uncertainty estimates quantitatively using the Sparsification Plot and the Error Plot, and qualitatively by showing the reconstructions with low and high uncertainty estimates. As shown in the Error Plots, both visual and lidar depth uncertainty can indicate the degree of the reconstruction error. In particular, we can observe how the visual uncertainty and lidar depth uncertainty capture different parts of the scene according to the sensor characteristics.
From the reconstruction error figure of both CARLA and ROQ, it can be seen that reconstructions with low visual uncertainty generally are places where visual features can be detected. In fact, this corresponds to the visual features that can be reliably estimated by classical SfM and visual SLAM methods. In ROQ, much of the ground in the quad has relatively higher visual uncertainty but lower lidar depth uncertainty. Essentially, even if there are few visual features on the ground which are not ideal for visual reconstruction, the lidar measurements provide sufficient information to produce an accurate ground reconstruction. When uncertainty estimates are high, we found that lidar depth uncertainty is a better indicator of the reconstruction error than visual uncertainty.  From the Error Plot, the average error of points with high lidar depth uncertainty estimates (blue curve) is generally higher than the error of points with high visual uncertainty estimates (red curve). This can also be shown in the Sparsification Plot: as the first 20\% of the reconstruction of higher uncertainty are being removed, the sparsification curve by depth uncertainty (blue curve) is closer to the ideal oracle sparsification (dashed blue curve) compared to the sparsification curve by visual uncertainty (red curve). This is because the depth uncertainty estimates remove points with higher errors than the visual uncertainty estimates, and hence the errors of the remaining points are lower.

\begin{figure}[h]
	\centering
	\includegraphics[width=1\linewidth]{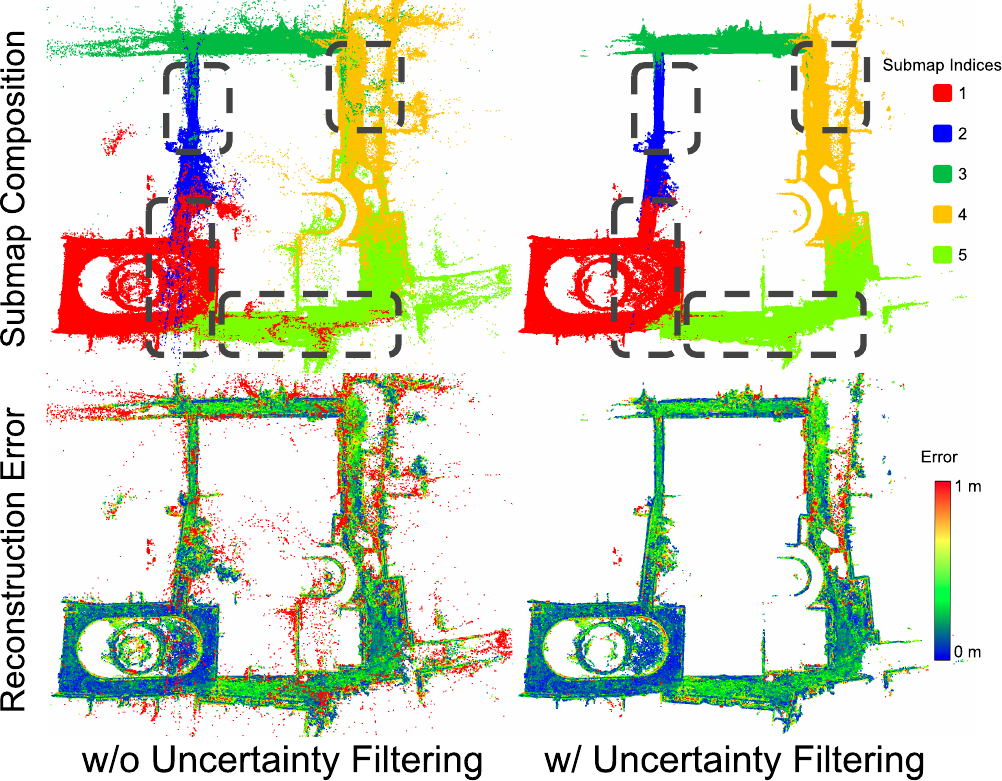}
	\caption{Comparison between two merged NeRF submaps with and without uncertainty filtering. On the left, the submaps contain artefacts which extend to neighbouring submaps and reduce overall reconstruction quality. These artefacts are mainly due to insufficient observations at the submap boundaries. On the right, we show how uncertainty filtering can be used to remove these artefacts which leads to a more accurate merged reconstruction.}
	\label{fig:submap_unc_merge}
\end{figure}

When merging NeRF submaps, uncertainty-based filtering is particularly important. As shown in \figref{fig:submap_unc_merge}, the submaps can contain reconstruction artefacts which interfere with neighbouring submaps due to limited observation at the submap boundary. These artefacts can be however identified as they tend to have high epistemic uncertainty, and can be filtered to improve the merged reconstruction accuracy. 

\begin{figure}[h]
	\centering
	\includegraphics[width=0.95\columnwidth]{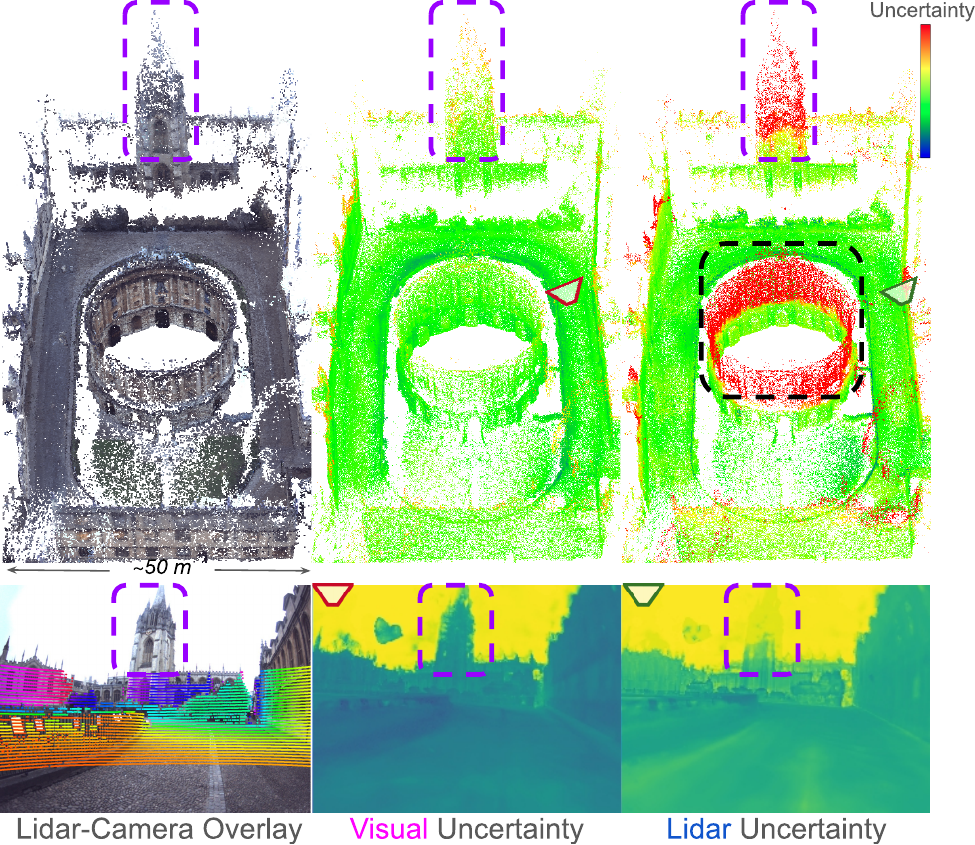}
	\caption{Comparison between visual and lidar uncertainty. The reconstructions at the top are coloured by uncertainty estimates (red is high uncertainty, green is low uncertainty). Here, the lidar used was a Hesai XT32 which has a narrow FOV and cannot scan the upper part of the buildings, and the depth uncertainty estimates can identify such regions (red point clouds). This indicates that only visual information is used when reconstructing these areas.}
	\label{fig:lidar_visual_unc}
\end{figure}

The advantage of the disentangled uncertainty estimates is also demonstrated in \figref{fig:lidar_visual_unc} where we use a narrow vertical FOV lidar with wider vertical FOV cameras. Here, the lidar is not able to scan the upper part of the two buildings highlighted in \figref{fig:lidar_visual_unc}, but the cameras can. Because of this, when we compute the epistemic uncertainty, we can see high lidar depth uncertainty for the upper part of both buildings. This indicates that the reconstruction is derived primarily from the visual data. In summary, from a sensor fusion point of view, our epistemic uncertainty estimation framework provides a systematic analysis of each sensor's contribution to the final reconstruction.

\subsection{View Selection Strategy}

\begin{figure}[h]
	\centering
	\includegraphics[width=\columnwidth]{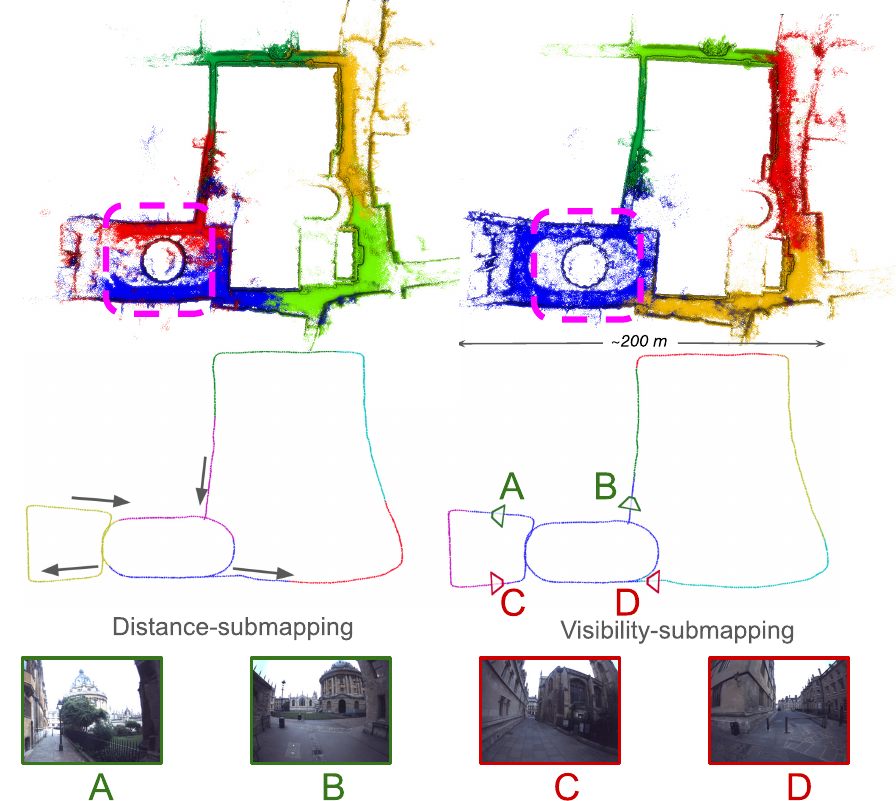}
	\caption{Comparison of two submapping strategies. Visibility information guides the clustering algorithm to group images looking at the same object together, which then leads to a more accurate and complete reconstruction of that object than algorithms that only consider distances.}
	\label{fig:submapping_compare}
\end{figure}

In this section, we compare our visibility-based submapping strategy with an alternative distance-based submapping strategy proposed in~\cite{tao2024silvr}. As shown in \figref{fig:submapping_compare}, the building highlighted is divided into two submaps when using the distance-based submapping method. This is not ideal as it reduces the number of view constraints, which makes each submap's partial reconstruction of that building have a lower quality. In addition, distance-based submapping puts the poses in A and C into the same submap, which is in fact not ideal. While pose A and pose C are spatially close, they have opposite viewing directions: pose A is looking at the highlighted building, while pose C is looking away from it. In comparison, visibility-based submapping moves pose A into the submap that contains the highlighted building, and pose C into another submap.








\subsection{Multi-Camera Setup Ablation Study}
\begin{figure}[h]
	\centering
	\includegraphics[width=0.98\columnwidth]{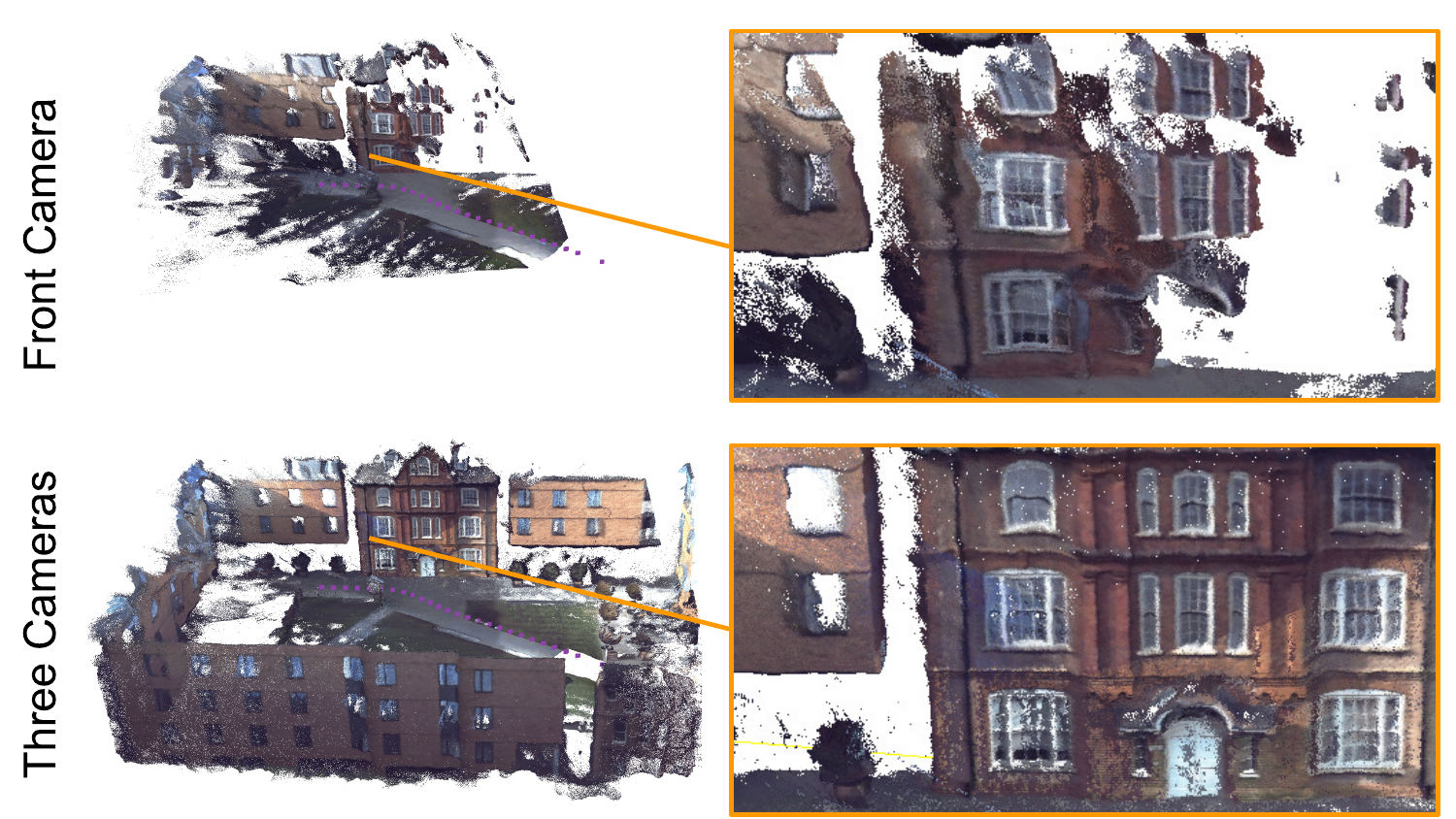}
	\caption{Comparison of reconstruction of HBAC building using the front camera only vs. using all three cameras. The three-camera setup generates more complete and accurate reconstructions compared to using only a single front-facing camera. The multi-camera setting is important in robotic applications where it would be infeasible to actively scan the entire scene to obtain strong viewpoint constraints.}
	\label{fig:mono_3cam}
\end{figure}

The advantage of our multi-camera sensor setup is demonstrated qualitatively in \figref{fig:mono_3cam}. Compared to the three-camera setup, using only the front-facing camera leads to a reconstruction that is not only incomplete, but also with poorer geometry. Visual reconstruction with the photometric loss is limited to generating good-quality rendering only at the input viewing angle. The reconstruction using the front-only camera in \figref{fig:mono_3cam} is trained with images looking in a single direction in the scene. This results in a poor geometric reconstruction when rendered from an unseen angle. In comparison, reconstruction with three cameras generates a more complete and more accurate reconstruction.




\subsection{Effect of Bootstrapping SLAM Poses}
We compare the performance of different strategies for computing poses: SLAM poses produced online, SLAM poses later refined using NeRF~\cite{nerfstudio}, SLAM poses refined using COLMAP~\cite{schoenberger2016colmap}'s Bundle Adjustment in different configurations, and COLMAP without any prior poses. For COLMAP, we tested different numbers of features extracted per image, as well as two different COLMAP feature matching algorithms: sequential matching with loop closures and Vocabulary Tree Matcher~\cite{schonberger2017vote}. 

The results are summarised in \tabref{tab:pose_ablation}. For all COLMAP configurations, providing the SLAM prior poses not only accelerates pose computation, but also leads to better test rendering, compared to COLMAP without any initialisation. Our SLAM prior poses can also register all the images in the trajectory; meanwhile, COLMAP on its own only registers only 55\%-95\% images. Extracting more visual features per image (from 1024 to 8192) leads to a higher percentage of image registration and better visual reconstruction (PSNR and SSIM). This comes at the expense of a higher computation time, especially with the VocabTree matcher. Using the COLMAP Sequential Matcher is generally faster than Vocabulary Tree Matcher.

\begin{table}[h]
	\caption{\small{Ablation: Effect of Bootstrapping w/ SLAM Poses}}
  	\setlength{\tabcolsep}{3pt} 
	\centering
 	\begin{tabular}{ l c c r r r r r}
		\toprule
		Method & Features & Prior& Traj. Regis  & \multicolumn{2}{c}{PSNR$\uparrow$} &SSIM$\uparrow$   & Time 
		\\
            &&& tered & Train&Test&Test
		\\
            &&&(\%)&&&&(s)
            \\
		\hline
		\addlinespace

            VILENS &/&/&100.0&23.0&17.4&0.64&Online
            \\
            NeRF refined &/&/&100.0&23.2&17.9&0.65&Online
            \\
            \addlinespace

            \multirow{4}{1cm}{COLMAP Sequential} & 1024&   &57.6&25.9&19.1&0.71&3299.2
            \\
            &1024 &\checkmark &100.0&26.2&\textbf{20.6}&\textbf{0.74}&1729.9
            \\
            &8192 &&94.0&26.1&19.8&0.72&7850.0
            \\
            &8192 &\checkmark &100.0&26.2&20.4&0.73&4448.4
            \\
		\addlinespace
            \multirow{4}{1cm}{COLMAP VocabTree} & 1024&   & 54.7&26.2&19.0&0.71&4444.8
            \\
            &1024 &\checkmark & 100.0&26.3&20.4&0.73& 1052.5
            \\
            &8192  &&94.8&\textbf{26.6}&19.9&0.72&37067.5
            \\
            &8192 &\checkmark &100.0&26.3&20.4&\textbf{0.74}&11015.3
            \\
		\bottomrule
		\addlinespace
	\end{tabular}
	\footnotesize{Results evaluated on HBAC-Maths dataset with 3254 images and duration of 1270s. Models trained for 4000 iterations. 
 PSNR and SSIM were evaluated after masking out the sky.}
	\label{tab:pose_ablation}
\end{table}

\section{Conclusions}

In summary, we proposed a large-scale 3D reconstruction system fusing both lidar and vision in a neural radiance field. The proposed approach combines the advantages of the two sensor modalities and generates reconstructions with both photo-realistic textures as well as accurate geometry. We proposed a principled approach to quantification reconstruction uncertainty considering each sensor's characteristics, which enables us to identify unreliable reconstruction artefacts and filter them out to improve reconstruction accuracy. With our proposed submapping approach, we demonstrate large-scale reconstruction results from real-world datasets collected in different robot platforms in conditions suited to industrial inspection tasks.

\section*{Acknowledgements}
The authors would like to thank Matias Mattamala for discussion, proofreading and helping with figures.

\bibliographystyle{./IEEEtran}
\bibliography{./IEEEabrv,./library}

\vskip 0pt plus -1fil

\begin{IEEEbiography}[{\includegraphics[width=1in,height=1.25in,clip,keepaspectratio]{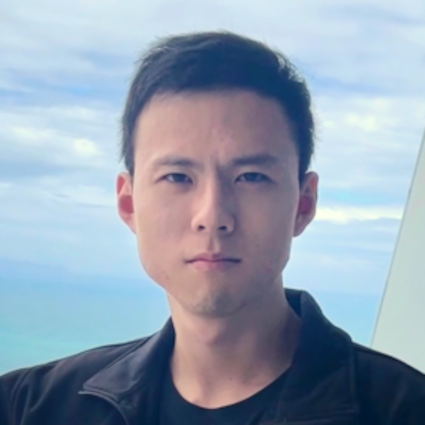}}]{Yifu Tao}
(Graduate Student Member, IEEE) 
received an M.Eng. degree in Engineering Science from the University of Oxford, UK, in 2020. 
He received the DPhil degree in Engineering Science from the University of Oxford, UK, in 2025.
He is currently a postdoctoral researcher in the Oxford Robotics Institute at the University of Oxford.
His research interests include 3D reconstruction using visual and lidar sensors and deep learning methods.
\end{IEEEbiography}

\vskip 0pt plus -1fil

\begin{IEEEbiography}[{\includegraphics[width=1in,height=1.25in,clip,keepaspectratio]{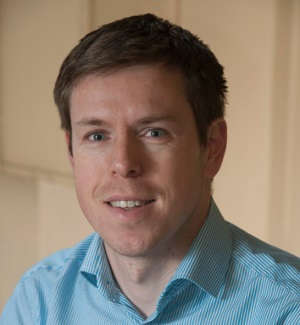}}]{Maurice Fallon}
(Senior Member, IEEE) received the
B.Eng. degree in electrical engineering from University College Dublin, Dublin, Ireland, in 2004 and the Ph.D. degree in acoustic source tracking from the University of Cambridge, Cambridge, U.K., in 2008.
From 2008 to 2012, he was a Postdoc and a Research Scientist with MIT Marine Robotics Group
working on SLAM. Later, he was the Perception Lead of MIT’s team in the DARPA Robotics Challenge.
Since 2017, he has been a Royal Society University Research Fellow and an Associate Professor with the University of Oxford, Oxford, U.K. He leads the Dynamic Robot Systems Group,
Oxford Robotics Institute. His research interests include probabilistic methods
for localization, mapping, multisensor fusion, and robot navigation. His research has won or been nominated for best paper awards at 6 IEEE conferences (ICRA, Humanoids and IV).
\end{IEEEbiography}

\end{document}